\documentclass[10pt,twocolumn,letterpaper]{article}

\usepackage[utf8]{inputenc}
\usepackage{cvpr}
\usepackage{times}
\usepackage{epsfig}
\usepackage{graphicx}
\usepackage{amsmath,amsfonts,amssymb,amsthm}
\usepackage{algorithm}
\usepackage[noend]{algpseudocode}
\usepackage{color}
\usepackage{caption}
\usepackage{subcaption}
\usepackage{xspace}
\usepackage{array}
\usepackage{cite}
\usepackage[pagebackref=true,breaklinks=true,colorlinks,bookmarks=false]{hyperref}

\newcommand{\bS}{\mathbf{S}}

\newcommand{\bo}{\mathbf{o}}

\newcommand{\nE}{\mathbb{E}}

\newcommand{\cR}{\mathcal{R}}

\newcommand{\cC}{\mathcal{C}}

\newcommand{\cX}{\mathcal{X}}

\newcommand{\figref}[1]{\Fig~\ref{#1}}
\newcommand{\secref}[1]{Section~\ref{#1}}

\newcommand{\eqnref}[1]{Eq.~\eqref{#1}}
\newcommand{\tabref}[1]{Table~\ref{#1}}

\makeatletter
\DeclareRobustCommand\onedot{\futurelet\@let@token\@onedot}
\def\@onedot{\ifx\@let@token.\else.\null\fi\xspace}
\def\eg{e.g\onedot} 
\def\ie{i.e\onedot} 
\def\cf{cf\onedot}

\def\etal{et~al\onedot} 
\def\Fig{Fig\onedot}   
\makeatother

\newcommand{\boldparagraph}[1]{\vspace{0.2cm}\noindent{\bf #1:} }

\definecolor{darkgreen}{rgb}{0,0.7,0}

\newcolumntype{C}[1]{>{\centering\arraybackslash}p{#1}}

\cvprfinalcopy
\def\cvprPaperID{1719}

\ifcvprfinal\fi

\begin{document}

\title{RayNet: Learning Volumetric 3D Reconstruction with Ray Potentials}

\author{Despoina Paschalidou$^{1,5}$ \quad Ali Osman Ulusoy$^{2}$ \quad Carolin Schmitt$^{1}$  \quad Luc van Gool$^{3}$ \quad Andreas Geiger$^{1,4}$\\
$^1$Autonomous Vision Group, MPI for Intelligent Systems T{\"u}bingen\\
$^2$Microsoft \quad
$^3$Computer Vision Lab, ETH Z{\"u}rich \& KU Leuven\\
$^4$Computer Vision and Geometry Group, ETH Z{\"u}rich\\
$^5$Max Planck ETH Center for Learning Systems\\
{\tt\small \{firstname.lastname\}@tue.mpg.de \quad vangool@vision.ee.ethz.ch}}

\maketitle

\begin{abstract}
In this paper, we consider the problem of reconstructing a dense 3D model using images captured from different views. Recent methods based on convolutional neural networks (CNN) allow learning the entire task from data. However, they do not incorporate the physics of image formation such as perspective geometry and occlusion. Instead, classical approaches based on Markov Random Fields (MRF) with ray-potentials explicitly model these physical processes, but they cannot cope with large surface appearance variations across different viewpoints. In this paper, we propose RayNet, which combines the strengths of both frameworks. RayNet integrates a CNN that learns view-invariant feature representations with an MRF that explicitly encodes the physics of perspective projection and occlusion. We train RayNet end-to-end using empirical risk minimization. We thoroughly evaluate our approach on challenging real-world datasets and demonstrate its benefits over a piece-wise trained baseline, hand-crafted models as well as other learning-based approaches.
\end{abstract}

\section{Introduction}
\label{sec:introduction}

\begin{figure}
    \begin{subfigure}{\linewidth}
        \centering
        \includegraphics[width=0.9\textwidth]{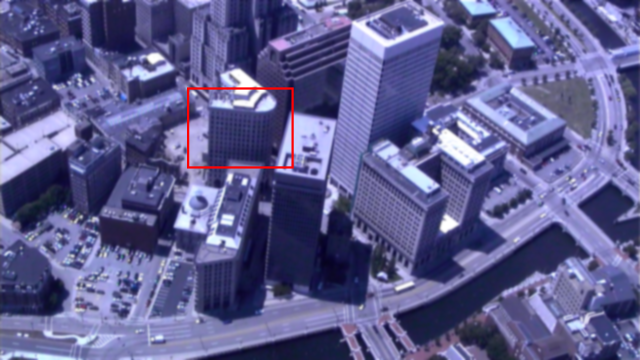} 
        \vspace{-0.2cm}   
        \caption{Input Image} 
        \vspace{0.1cm}        
    \end{subfigure}\\
    \begin{subfigure}{0.5\linewidth}
        \centering
        \includegraphics[width=0.8\linewidth]{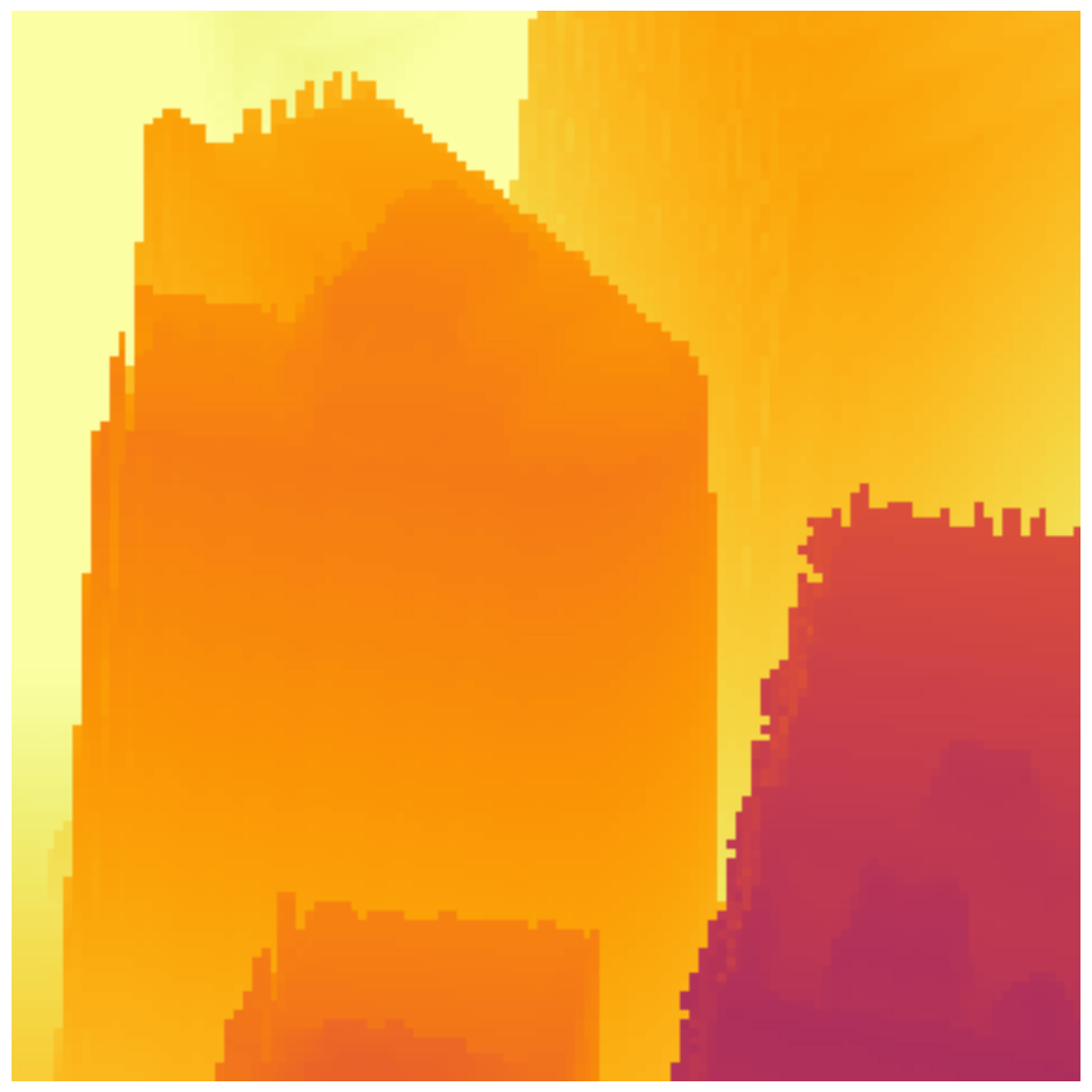} 
        \vspace{-0.2cm}
        \caption{Ground-truth} 
    \end{subfigure}
    \begin{subfigure}{0.5\linewidth}
        \centering
        \includegraphics[width=0.8\textwidth]{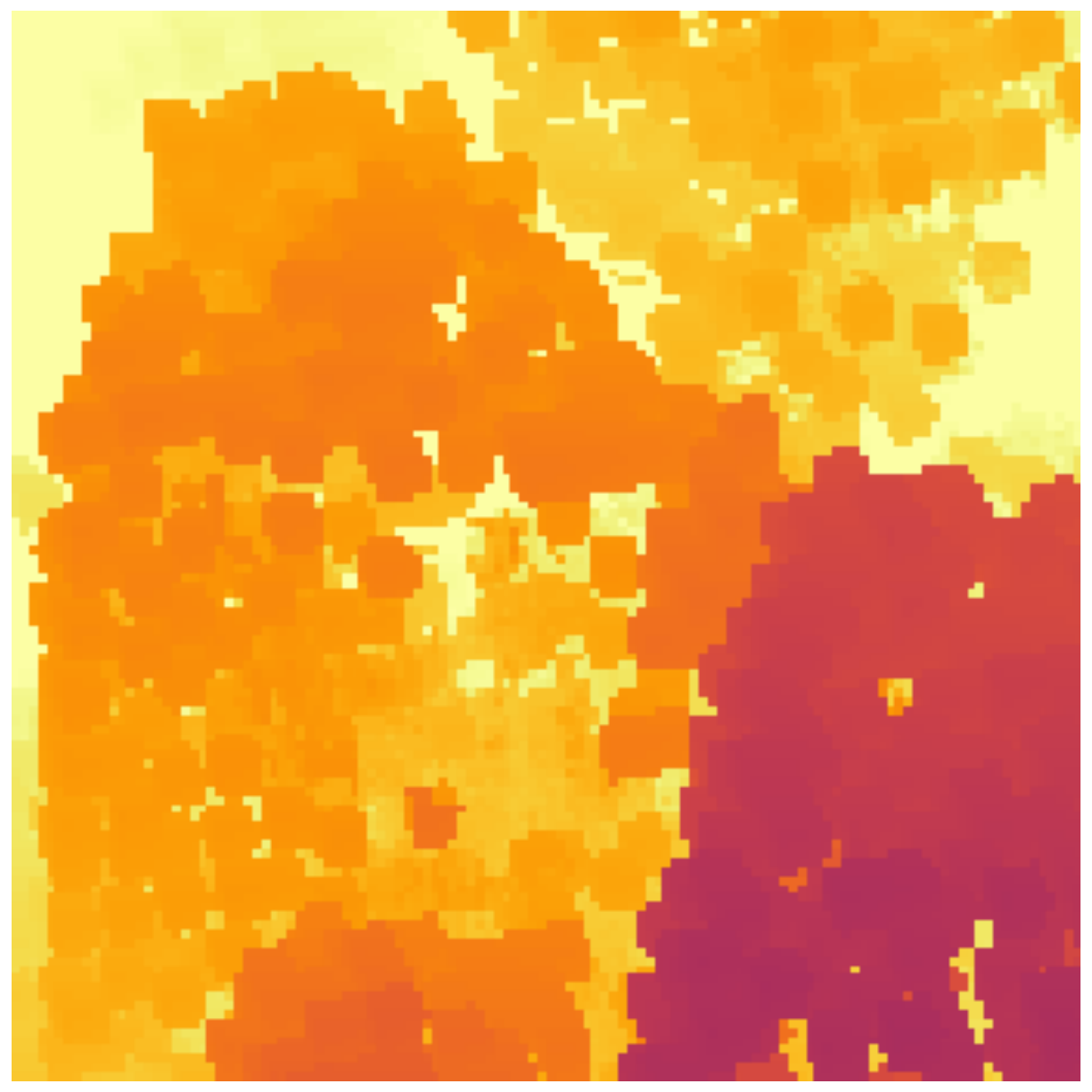} 
        \vspace{-0.2cm}   
        \caption{Ulusoy et al. \cite{Ulusoy2015THREEDV}} 
    \end{subfigure} 
    \begin{subfigure}{0.5\linewidth}
        \centering
        \includegraphics[width=0.8\textwidth]{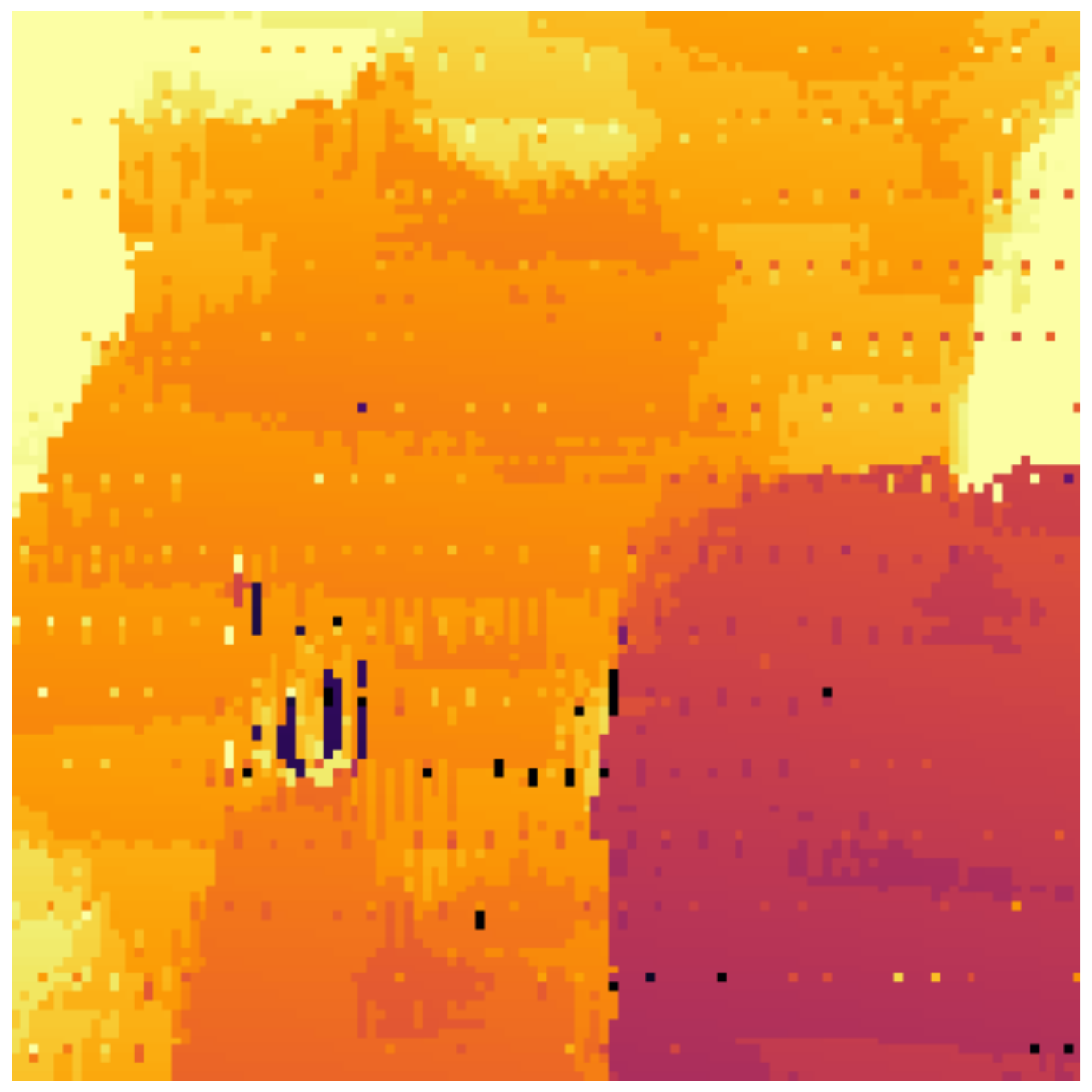} 
        \vspace{-0.2cm}   
        \caption{Hartmann et al. \cite{Hartmann2017ICCV}} 
    \end{subfigure}
    \begin{subfigure}{0.5\linewidth}
        \centering
        \includegraphics[width=0.8\textwidth]{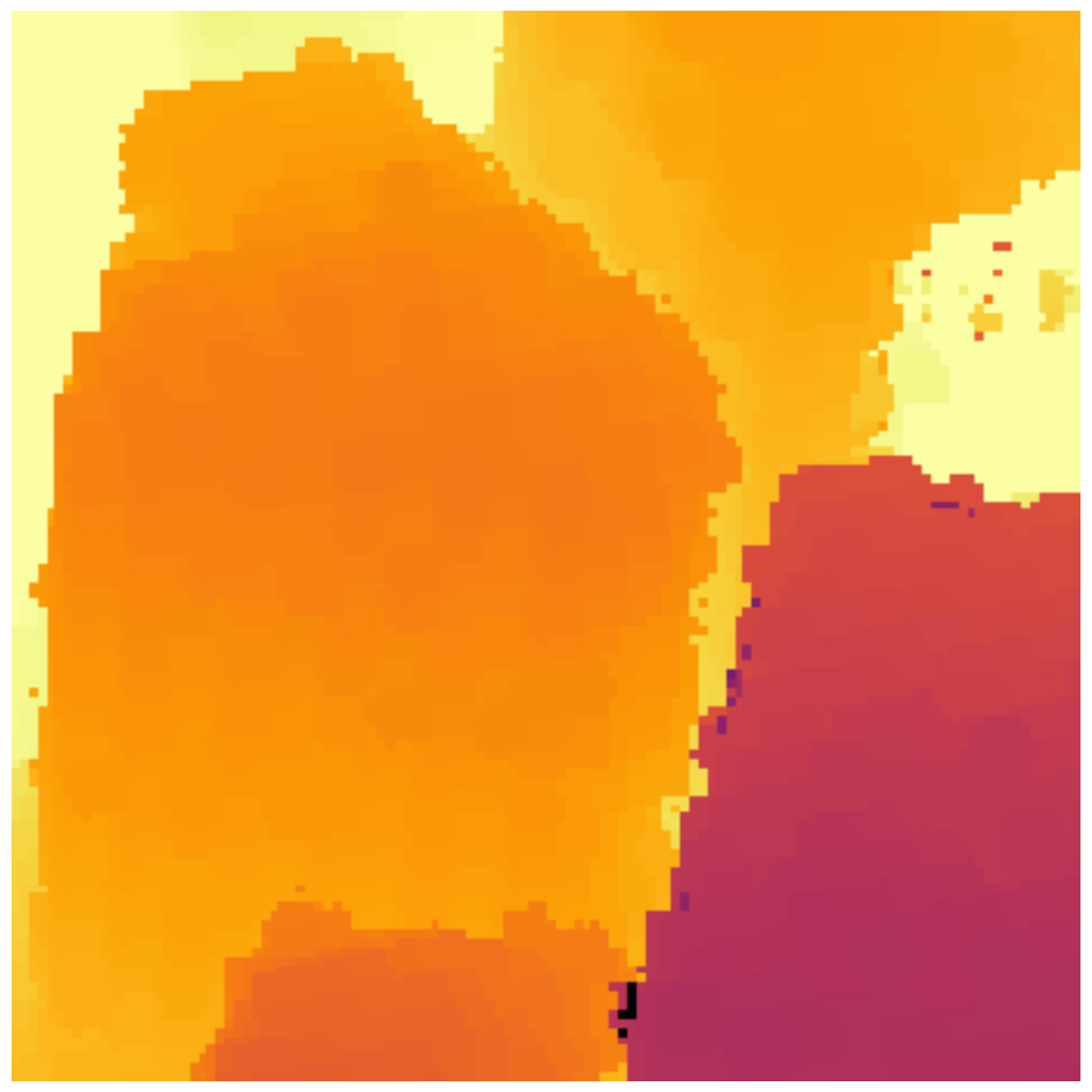} 
        \vspace{-0.2cm}   
        \caption{RayNet} 
    \end{subfigure}
    \caption{{\bf Multi-View 3D Reconstruction}. By combining representation learning with explicit physical constraints about perspective geometry and multi-view occlusion relationships, our approach (e) produces more accurate results than entirely model-based (c) or learning-based methods that ignore such physical constraints (d).
}
    \label{fig:teaser} 
    \vspace{-1.5em}
\end{figure}

Passive 3D reconstruction is the task of estimating a 3D model from a collection of 2D images taken from different viewpoints.
This is a highly ill-posed problem due to large ambiguities introduced by {\it  
occlusions} and {\it surface appearance variations} across different views.

Several recent works have approached this problem by formulating the task as inference in a Markov random field (MRF) with high-order ray potentials that explicitly model the physics of the image formation process along each viewing ray~\cite{Ulusoy2015THREEDV, Ulusoy2016CVPR, Liu2014PAMI}. The ray potential encourages consistency between the pixel recorded by the camera and the color of the first visible surface along the ray. By accumulating these constrains from each input camera ray, these approaches estimate a 3D model that is globally consistent in terms of occlusion relationships.

While this formulation correctly models occlusion, the complex nature of inference in ray potentials restricts these models to pixel-wise color comparisons, which leads to large ambiguities in the reconstruction~\cite{Ulusoy2015THREEDV}. Instead of using images as input, Savinov \etal~\cite{Savinov2015CVPR} utilize pre-computed depth maps using zero-mean normalized cross-correlation in a small image neighborhood. In this case, the ray potentials encourage consistency between the input depth map and the depth of the first visible voxel along the ray. While considering a large image neighborhood improves upon pixel-wise comparisons, our experiments show that such hand-crafted image similarity measures cannot handle complex variations of surface appearance. 

In contrast, recent learning-based solutions to motion estimation \cite{Ilg2017CVPR,Ranjan2017CVPR,Guney2016ACCV}, stereo matching \cite{Mayer2016CVPR,Zbontar2016JMLR,Luo2016CVPR} and 3D reconstruction \cite{Choy2016ECCV, Wu2016NIPS, Fan2017CVPR, Ji2017ICCV, Gwak2017ARXIV} have demonstrated impressive results by learning feature representations that are much more robust to local viewpoint and lighting changes. However, existing methods exploit neither the physical constraints of perspective geometry nor the resulting occlusion relationships across viewpoints, and therefore require a large model capacity as well as an enormous amount of labelled training data. 

This work aims at combining the benefits of a learning-based approach with the strengths of a model that incorporates the physical process of perspective projection and occlusion relationships.
Towards this goal, we propose an end-to-end trainable architecture called RayNet which integrates a convolutional neural network (CNN) that learns surface appearance variations (\eg across different viewpoints and lighting conditions) with an MRF that explicitly encodes the physics of perspective projection and occlusion.
More specifically, RayNet uses a learned feature representation that is correlated with nearby images to estimate surface probabilities along each ray of the input image set.
These surface probabilities are then fused using an MRF with high-order ray potentials that aggregates occlusion constraints across all viewpoints.
RayNet is learned end-to-end using empirical risk minimization. In particular, errors are backpropagated to the CNN based on the output of the MRF. This allows the CNN to specialize its representation to the joint task while explicitly considering the 3D fusion process.

Unfortunately, na\"{i}ve backpropagation through the unrolled MRF is intractable due to the large number of messages that need to be stored during training. We propose a stochastic ray sampling approach which allows efficient backpropagation of errors to the CNN.
We show that the MRF acts as an effective regularizer and improves both the output of the CNN as well as the output of the joint model for challenging real-world reconstruction problems.
Compared to existing MRF-based~\cite{Ulusoy2015THREEDV} or learning-based methods~\cite{Ji2017ICCV,Hartmann2017ICCV}, RayNet improves the accuracy of the 3D reconstruction by taking into consideration both local information around every pixel (via the CNN) as well as global information about the entire scene (via the MRF). 

Our code and data is available on the project website\footnote{\url{https://avg.is.tue.mpg.de/research_projects/raynet}}.

\section{Related Work}
\label{sec:related}
3D reconstruction methods can be roughly categorized into model-based and learning-based approaches, which learn the task from data.
As a thorough survey on 3D reconstruction techniques is beyond the scope of this paper, we discuss only the most related approaches and refer to \cite{Hartley2004,Furukawa2009ICCV, Seitz2006CVPR} for a more thorough review.

\boldparagraph{Ray-based 3D Reconstruction}
Pollard and Mundy \cite{Pollard2007CVPR} propose a volumetric reconstruction method that updates the occupancy and color of each voxel sequentially for every image. However, their method lacks a global probabilistic formulation. To address this limitation, a number of approaches have phrased 3D reconstruction as inference in a Markov random field (MRF) by exploiting the special characteristics of high-order ray potentials \cite{Ulusoy2015THREEDV, Ulusoy2016CVPR, Savinov2015CVPR, Liu2014PAMI}. Ray potentials allow for accurately describing the image formation process, yielding 3D reconstructions consistent with the input images.
Recently, Ulusoy \etal \cite{Ulusoy2017CVPR} integrated scene specific 3D shape knowledge  to further improve the quality of the 3D reconstructions.
A drawback of these techniques is that very simplistic photometric terms are needed to keep inference tractable, \eg, pixel-wise color consistency, limiting their performance.

In this work, we integrate such a ray-based MRF with a CNN that learns multi-view patch similarity. This results in an end-to-end trainable model that is more robust to appearance changes due to viewpoint variations, while tightly integrating perspective geometry and occlusion relationships across viewpoints. 

\begin{figure*}[t!]
	\begin{subfigure}[b]{0.75\linewidth}
		\centering
		\includegraphics[width=\linewidth]{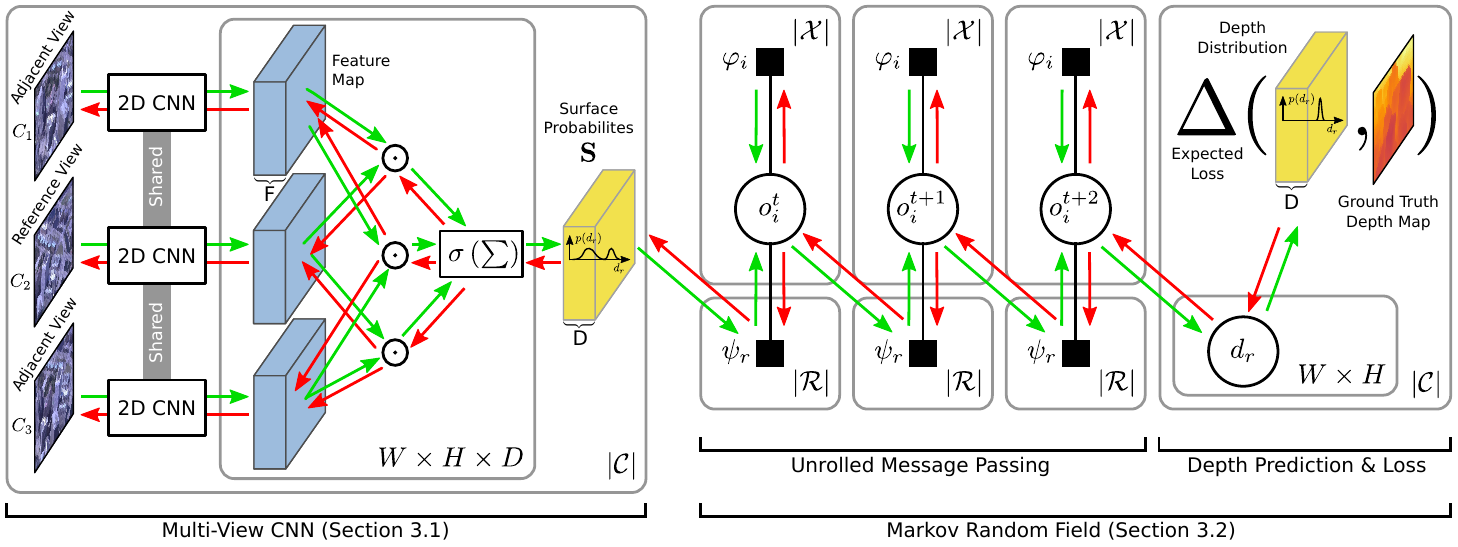}
		\caption{{\bf RayNet Architecture in Plate Notation with Plates denoting Copies}}
		\label{fig:network_architecture}
	\end{subfigure}\hfill
	\begin{subfigure}[b]{0.22\linewidth}
		\centering
		\raisebox{0.7cm}{\includegraphics[width=0.9\linewidth]{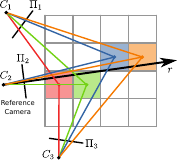}}
		\caption{{\bf Multi-View Projection}}
		\label{fig:network_multi_view}
	\end{subfigure}
	\caption{{\bf RayNet.} Given a reference view and its adjacent views, we extract features via a 2D CNN (\textcolor{blue}{blue}). Features corresponding to the projection of the same voxel along ray $r$ (see (\subref{fig:network_multi_view}) for an illustration) are aggregated via the average inner product into per pixel depth distributions. The average runs over all pairs of views and $\sigma(\cdot)$ denotes the softmax operator. The depth distributions for all rays (\ie, all pixels in all views) are passed to the unrolled MRF. The final depth predictions $d_r$ are passed to the loss function. The forward pass is illustrated in \textcolor{darkgreen}{green}. The backpropagation pass is highlighted in \textcolor{red}{red}. $\cC=\{C_i\}$ is the set of all cameras, $W\times H$ are the image dimensions and $D$ is the max. number of voxels along each ray.
	}
	\label{fig:network}
\end{figure*}

\boldparagraph{Learning-based 3D Reconstruction}
The development of large 3D shape databases \cite{Chang2015ARXIV,Wu2015CVPR} has fostered the development of learning based solutions \cite{Choy2016ECCV,Wu2016NIPS,Girdhar2016ECCV,Achlioptas2017ARXIV} to  3D reconstruction.
Choy \etal \cite{Choy2016ECCV} propose a unified framework for single and multi-view reconstruction by using a 3D recurrent neural network (3D-R2N2) based on long-short-term memory (LSTM).
Girdhar \etal  \cite{Girdhar2016ECCV} propose a network that embeds image and shape together for single view 3D volumetric shape generation.
Hierarchical space partitioning structures (\eg, octrees) have been proposed to increase the output resolution beyond $32^3$ voxels  \cite{Riegler2017THREEDV,Haene2017ARXIV,Tatarchenko2017ICCV}.

As most of the aforementioned methods solve the 3D reconstruction problem via recognizing the scene content, they are only applicable to object reconstruction and do not generalize well to novel object categories or full scenes. Towards a more general learning based model for 3D reconstruction, \cite{Kar2017NIPS,Ji2017ICCV} propose to unproject the input images into 3D voxel space and process the concatenated unprojected volumes using a 3D convolutional neural network. While these approaches take projective geometry into account, they do not explicitly exploit occlusion relationships across viewpoints, as proposed in this paper.
Instead, they rely on a generic 3D CNN to learn this mapping from data.
We compare to \cite{Ji2017ICCV} in our experimental evaluation and obtain more accurate 3D reconstructions and significantly better runtime performance. In addition, the lightweight nature of our model's forward inference allows for reconstructing scenes up to $256^3$ voxels resolution in a single pass. In contrast, \cite{Kar2017NIPS} is limited to $32^3$ voxels and \cite{Ji2017ICCV} requires processing large volumes using a sliding window, thereby losing global spatial relationships.

A major limitation of all aforementioned approaches is that they require full 3D supervision for training, which is quite restrictive.
Tulsiani \etal \cite{Tulsiani2017CVPR} relax these assumptions by formulating a differentiable view consistency loss that measures the inconsistency between the predicted 3D shape and its observation.
Similarly, Rezende \etal \cite{Rezende2016NIPS} propose a neural projection layer and a black box renderer for supervising the learning process.
Yan \etal \cite{Yan2016NIPS} and Gwak \etal \cite{Gwak2017ARXIV} use 2D silhouettes as supervision for 3D reconstruction from a single image.
While all these methods exploit ray constraints inside the loss function, our goal is to directly integrate the physical properties of the image formation process into the model via unrolled MRF inference with ray potentials.
Thus, we are able to significantly reduce the number of parameters in the network and our network does not need to acquire these first principles from data.

\section{Model}
\label{sec:method}
The input to our approach is a set of images and their corresponding camera poses, which are obtained using structure-from-motion \cite{Wu2011CVPR}. 
Our goal is to model the known physical processes of perspective projection and occlusion, while learning the parameters that are difficult to model, \eg, those describing surface appearance variations across different views. 
Our architecture utilizes a CNN to learn a feature representation for image patches that are compared across nearby views to estimate a depth distribution for each ray in the input images. For all our experiments, we use $4$ nearby views. Due to the small size of each image patch, these distributions are typically noisy. We pass these noisy distributions to our MRF which aggregates them into a occlusion-consistent 3D reconstruction. We formulate inference in the MRF as a differentiable function, hence allowing end-to-end training using backpropagation. 

We first specify our CNN architecture which predicts depth distributions for each input pixel/ray. We then detail the MRF for fusing these noisy measurements. Finally, we discuss appropriate loss functions and show how our model can be efficiently trained using stochastic ray sampling.
The overall architecture is illustrated in \figref{fig:network}.

\subsection{Multi-View CNN}
\label{sec:cnn}

CNNs have been proven successful for learning similarity measures between two or more image patches for stereo~\cite{Zagoruyko2015CVPR, Zbontar2014ARXIV, Luo2016CVPR} as well as multi-view stereo~\cite{Hartmann2017ICCV}.
Similar to these works, we design a network architecture that estimates a depth distribution for every pixel in each view.

The network used is illustrated in Fig.~\ref{fig:network_architecture} (left).
The network first extracts a $32$-dimensional feature per pixel in each input image using a fully convolutional network. The weights of this network are shared across all images. Each layer comprises convolution, spatial batch normalization and a ReLU non-linearity. We follow common practice and remove the ReLU from the last layer in order to retain information encoded both in the negative and positive range~\cite{Luo2016CVPR}.

We then use these features to calculate per ray depth distributions for all input images.
More formally, let $\cX$ denote a voxel grid which discretizes the 3D space. Let $\cR$ denote the complete set of rays in the input views. In particular, we assume one ray per image pixel. Thus, the cardinality of $\cR$ equals the number of input images times the number of pixels per image. For every voxel $i$ along each ray $r \in \cR$, we compute the surface probability $s_i^r$ by projecting the voxel center location into the reference view (\ie, the image which comprises pixel $r$) and into all adjacent views ($4$ views in our experiments)
as illustrated in \figref{fig:network_multi_view}. For clarity, \figref{fig:network_multi_view} shows only $2$ out of $4$ adjacent cameras considered in our experimental evaluation.
We obtain $s_i^r$ as the average inner product between all pairs of views. Note that $s_i^r$ is high if all views agree in terms of the learned feature representation. Thus, $s_i^r$ reflects the probability of a surface being located at voxel $i$ along ray $r$. We abbreviate the surface probabilities for all rays of a single image with $\bS$.

\subsection{Markov Random Field}
\label{sec:mrf}

Due to the local nature of the extracted features as well as occlusions, the depth distributions computed by the CNN are typically noisy.
Our MRF aggregates these depth distributions by exploiting occlusion relationships across all viewpoints, yielding significantly improved depth predictions.
Occlusion constraints are encoded using high-order ray potentials \cite{Ulusoy2015THREEDV,Savinov2015CVPR,Liu2014PAMI}. We differ from \cite{Ulusoy2015THREEDV,Liu2014PAMI} in that we do not reason about surface appearance within the MRF but instead incorporate depth distributions estimated by our CNN. This allows for a more accurate depth signal and also avoids the costly sampling-based discrete-continuous inference in \cite{Ulusoy2015THREEDV}. We compare against \cite{Ulusoy2015THREEDV} in our experimental evaluation and demonstrate improved performance. 

We associate each voxel $i \in \cX$ with a binary occupancy variable $o_i \in \{0, 1\}$, indicating whether the voxel is occupied ($o_i=1$) or free ($o_i=0$).
For a single ray $r \in \cR$, let $\bo_r = (o_1^r, o_2^r,\dots, o_{N_r}^r)$ denote the ordered set of occupancy variables associated with the voxels which intersect ray $r$.  The order is defined by the distance to the camera. 

\begin{figure}[t!]
	\centering
	\includegraphics[width=\linewidth]{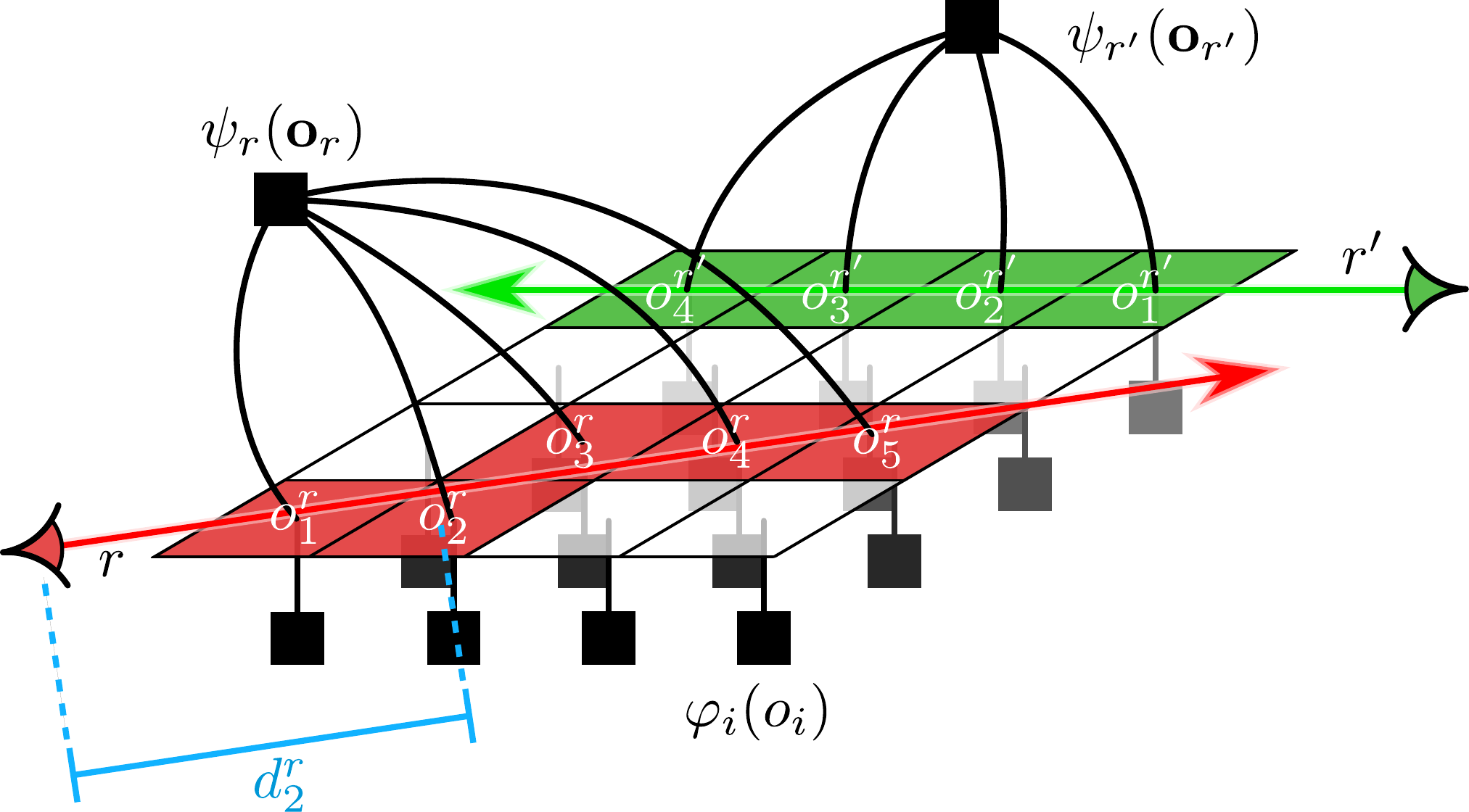}
	\caption{{\bf Factor Graph of the MRF.}
		We illustrate a 2D slice through the 3D volume and two cameras with one ray each: $r,r'\in\cR$. Unary factors $\varphi_i(o_i)$ are defined for every voxel $i\in\cX$ of the voxel grid. Ray factors $\psi_r(\bo_r)$ are defined for every ray $r\in\cR$ and connect to the variables along the ray $\bo_r = (o_1^r,\dots,o_{N_r}^r)$. $d_i^r$ denotes the Euclidean distance of the $i$th voxel along ray $r$ to its corresponding camera center.}
	\label{fig:mrf}
	\vspace{-0.3em}
\end{figure}

The joint distribution over all occupancy variables $\bo=\{o_i|i\in\cX\}$ factorizes into unary and ray potentials
\begin{equation}
	\begin{aligned}
		p(\bo) = \frac{1}{Z} \prod_{i \in \cX}
		\underbrace{\varphi_i(o_i)}_{\text{unary}}
		\prod_{r \in \cR}\underbrace{\psi_r(\bo_r)}_{\text{ray}}
		\label{eq:joint_probability}
	\end{aligned}
\end{equation}
where Z denotes the partition function. The corresponding factor graph is illustrated in \figref{fig:mrf}.

\boldparagraph{Unary Potentials} The unary potentials encode our prior belief about the state of the occupancy variables. We model $\varphi_i(o_i)$ using a Bernoulli distribution
\begin{equation}
	\varphi_i(o_i) = \gamma^{o_i}(1 - \gamma)^{1-o_i}
\end{equation}
where $\gamma$ is the probability that the $i$th voxel is occupied.

\boldparagraph{Ray Potentials} The ray potentials encourage the predicted depth at pixel/ray $r$ to coincide with the first occupied voxel along the ray. More formally, we have
\begin{equation}
	\psi_r(\mathbf{o_r}) = \sum_{i=1}^{N_r} o_i^r \prod_{j<i} (1-o_j^r)s_i^r\\
	\label{eq:depth_ray_potential}
\end{equation}
where $s_i^r$ is the probability that the visible surface along ray $r$ is located at voxel $i$. This probability is predicted by the neural network described in \secref{sec:cnn}. Note that the product over the occupancy variables in \eqnref{eq:depth_ray_potential} equals $1$ if and only if $i$ is the first occupied voxel (\ie, if $o_i^r=1$ and $o_j^r=0$ for $j<i$). Thus $\psi_r(\cdot)$ is large if the surface is predicted at the first occupied voxel in the model.

\boldparagraph{Inference}
Having specified our MRF, we detail our inference approach in this model. Provided noisy depth measurements from the CNN ($s_i^r$), the goal of the inference algorithm is to aggregate these measurements using ray-potentials into a 3D reconstruction and to estimate globally consistent depth distributions at every pixel. 

Let $d_i^r$ denote the distance of the $i$th voxel along ray $r$ to the respective camera center as illustrated in \figref{fig:mrf}.
Let further $d_r \in\{d_1^r,\dots,d_{N_r}^r\}$ be a random variable representing the depth along ray $r$. In other words, $d_r=d_i^r$ denotes that the depth at ray $r$ is the distance to the $i$th voxel along ray $r$. We associate the occupancy and depth variables along a ray using the following equation:
\begin{equation}
	d_r = \sum_{i=1}^{N_r} o_i^r \prod_{j<i} (1-o_i^r) d_i^r\\
	\label{eq:depth_occ_ray_potential}
\end{equation}
Our inference procedure estimates a probability distribution $p(d_r=d_i^r) $ for each pixel/ray $r$ in the input views. Unfortunately computing the exact solution in a loopy high-order model such as our MRF is NP-hard~\cite{Nowozin2011FTCGV}. We use loopy sum-product belief propagation for approximate inference. As demonstrated by Ulusoy \etal~\cite{Ulusoy2015THREEDV}, belief propagation in models involving high-order ray potentials is tractable as the factor-to-variable messages can be computed in linear time due to the special structure of the ray potentials.
In practice, we run a fixed number of iterations interleaving factor-to-variable and variable-to-factor message updates. We observed that convergence typically occurs after $3$ iterations and thus fix the iteration number to $3$ when unrolling the message passing algorithm.
We refer to the supplementary material for message equation derivations.

\subsection{Loss Function}
\label{sec:loss}

We utilize empirical risk minimization to train RayNet. Let $\Delta(d_r,d_r^\star)$ denote a loss function which measures the discrepancy between the predicted depth $d_r$ and the ground truth depth $d_r^\star$ at pixel $r$.
The most commonly used metric for evaluating depth maps is the absolute depth error. We therefore use the $\ell_1$ loss $\Delta(d_r,d_r^\star)=|d_r-d_r^\star|$ to train RayNet. In particular, we seek to minimize the expected loss, also referred to as the empirical risk $R(\theta)$:
\begin{eqnarray}
R(\theta) &=& \sum_{r\in\cR^{\star}} \nE_{p(d_r)} \, \Delta(d_r,d_r^\star)\label{eq:loss}\\
 &=& \sum_{r\in\cR^{\star}} \sum_{i=1}^{N_r} p(d_r=d_i^r) \, \Delta(d_i^r,d_r^\star)\nonumber
\end{eqnarray}
with respect to the model parameters $\theta$.
Here, $\cR^\star$ denotes the set of ground truth pixels/rays in all images of all training scenes and $p(d_r)$ is the depth distribution predicted by the model for ray $r$ of the respective image in the respective training scene. The parameters
$\theta$ comprises the occupancy prior $\gamma$ as well as parameters of the neural network.

\subsection{Training}

In order to train RayNet in an end-to-end fashion, we need to backpropagate the gradients through the unrolled MRF message passing algorithm to the CNN. However, a na\"{i}ve implementation of backpropagation is not feasible due to memory limitations. In particular, backpropagation requires storing {\it all} intermediate messages from {\it all} belief-propagation iterations in memory. For a modest dataset of $\approx$ 50 images with $360 \times 640$ pixel resolution and a voxel grid of size $256^3$, this would require $> 100$GB GPU memory, which is intractable using current hardware.

To tackle this problem, we perform backpropagation using mini-batches where each mini-batch is a stochastically sampled subset of the input rays. In particular, each mini-batch consists of $2000$ rays randomly sampled from a subset of 10 consecutive input images. Our experiments show that learning with rays from neighboring views leads to faster convergence, as the network can focus on small portion of the scene at a time.
After backpropagation, we update the model parameters and randomly select a new set of rays for the next iteration. This approximates the true gradient of the mini-batch.
The gradients are obtained using TensorFlow's AutoDiff functionality~\cite{Abadi2016ARXIV}.

While training RayNet in an end-to-end fashion is feasible, we further speed it up by pretraining the CNN followed by fine-tuning the entire RayNet architecture.
For pretraining, we randomly pick a set of pixels from a randomly chosen reference view for each mini-batch.
We discretize the ray corresponding to each pixel according to the voxel grid and project all intersected voxel centers into the adjacent views  as illustrated in \figref{fig:network_multi_view}. 
For backpropagation we use the same loss function as during end-to-end training, \cf Eq. \eqref{eq:loss}.

\section{Experimental Evaluation}
\label{sec:results}

In this Section, we present experiments evaluating our method on two challenging datasets.
The first dataset consists of two scenes, \textsc{BARUS\&HOLLEY} and \textsc{DOWNTOWN}, both of which were captured in urban environments from an aerial platform. The images, camera poses and LIDAR point cloud are provided by Restrepo et al. \cite{Restrepo2014JPRS}. Ulusoy et al. triangulated the LIDAR point cloud to achieve a dense ground truth mesh \cite{Ulusoy2015THREEDV}. In total the dataset consists of roughly 200 views with an image resolution of $1280 \times 720$ pixels.
We use the \textsc{BARUS\&HOLLEY} scene as the training set and  reserve \textsc{DOWNTOWN} for testing.

Our second dataset is the widely used DTU multi-view stereo benchmark~\cite{Aanes2016IJCV}, which comprises 124 indoor scenes of various objects captured from 49 camera views under seven different lighting conditions. We evaluate RayNet on two objects from this dataset: \textsc{BIRD} and \textsc{BUDDHA}.
For all datasets, we down-sample the images such that the largest dimension is 640 pixels.

We compare RayNet to various baselines both qualitatively and quantitatively. For the quantitative evaluation, we compute \emph{accuracy}, \emph{completeness} and \emph{per pixel mean depth error}, and report the mean and the median for each. The first two metrics are estimated in 3D space, while the latter is defined in image space and averaged over all ground truth depth maps. In addition, we also report the \emph{Chamfer distance}, which is a metric that expresses jointly the accuracy and the completeness.
Accuracy is measured as the distance from a point in the reconstructed point cloud to its closest neighbor in the ground truth point cloud, while completeness is calculated as the distance from a point in the ground truth point cloud to its closest neighbor in the predicted point cloud. For additional information on these metrics we refer the reader to \cite{Aanes2016IJCV}.
We generate the point clouds for the accuracy and completeness computation by projecting every pixel from every view into the 3D space according to the predicted depth and the provided camera poses.

\subsection{Aerial Dataset}
\label{sec:restrepo}

\begin{figure*}[t!]
	\centering
	\begin{subfigure}[b]{0.3\textwidth}
		\centering
		\includegraphics[width=\textwidth]{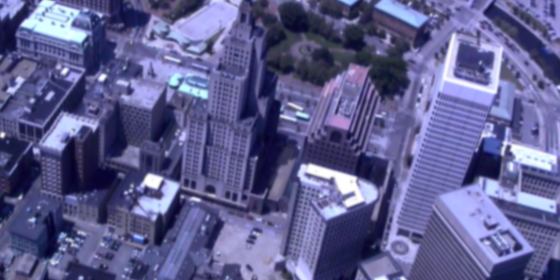}
		\caption[Image]%
		{{\small Image}}
		\label{fig:ground_truth}
	\end{subfigure}
	\hfill
	\begin{subfigure}[b]{0.3\textwidth}
		\centering
		\includegraphics[width=\textwidth]{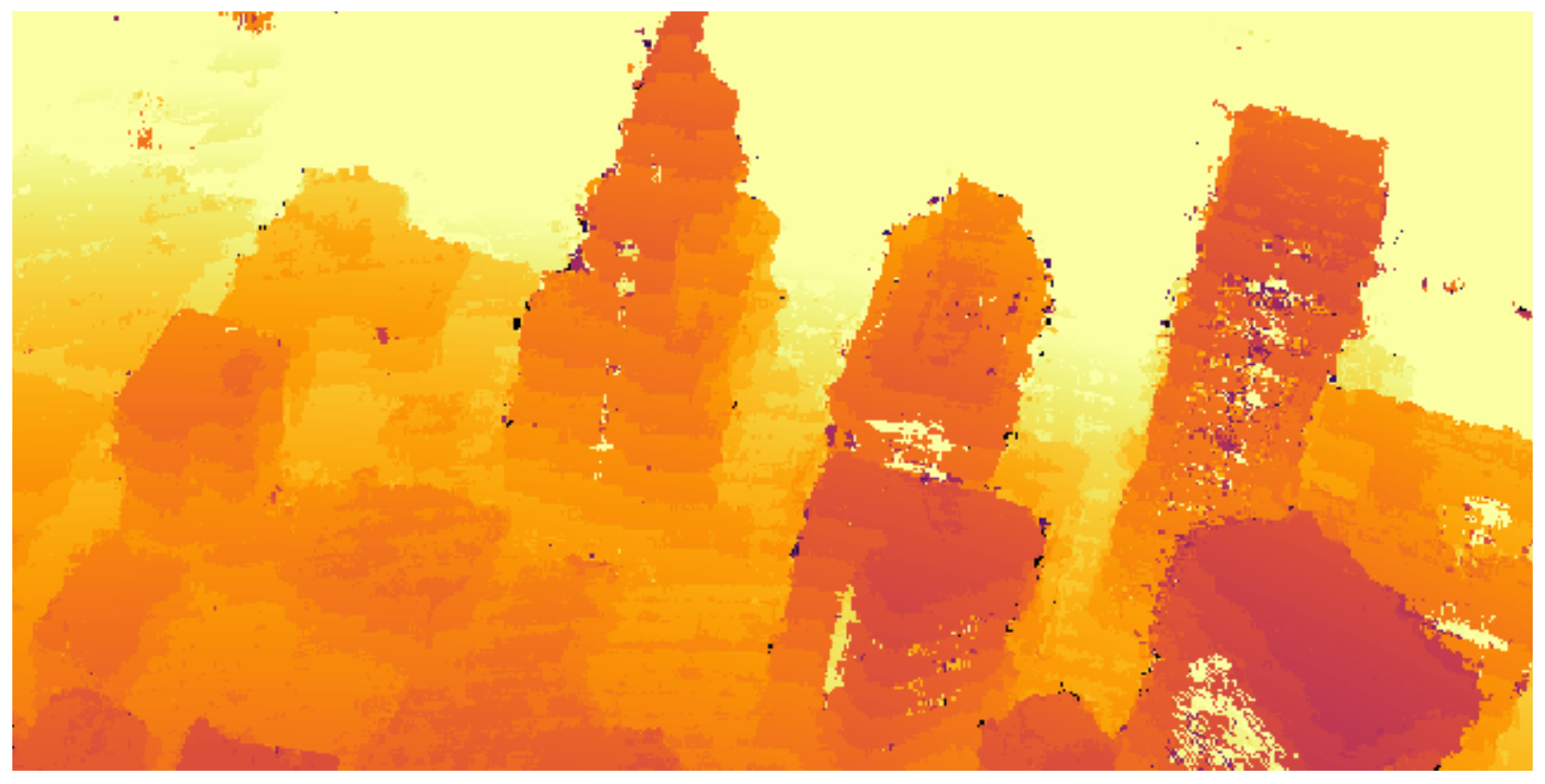}
		\caption[CNN pretrained]%
		{{\small Ours (CNN)}}    
		\label{fig:cnn_pretrained}
	\end{subfigure}
	\hfill
	\begin{subfigure}[b]{0.3\textwidth}
		\centering
		\includegraphics[width=\textwidth]{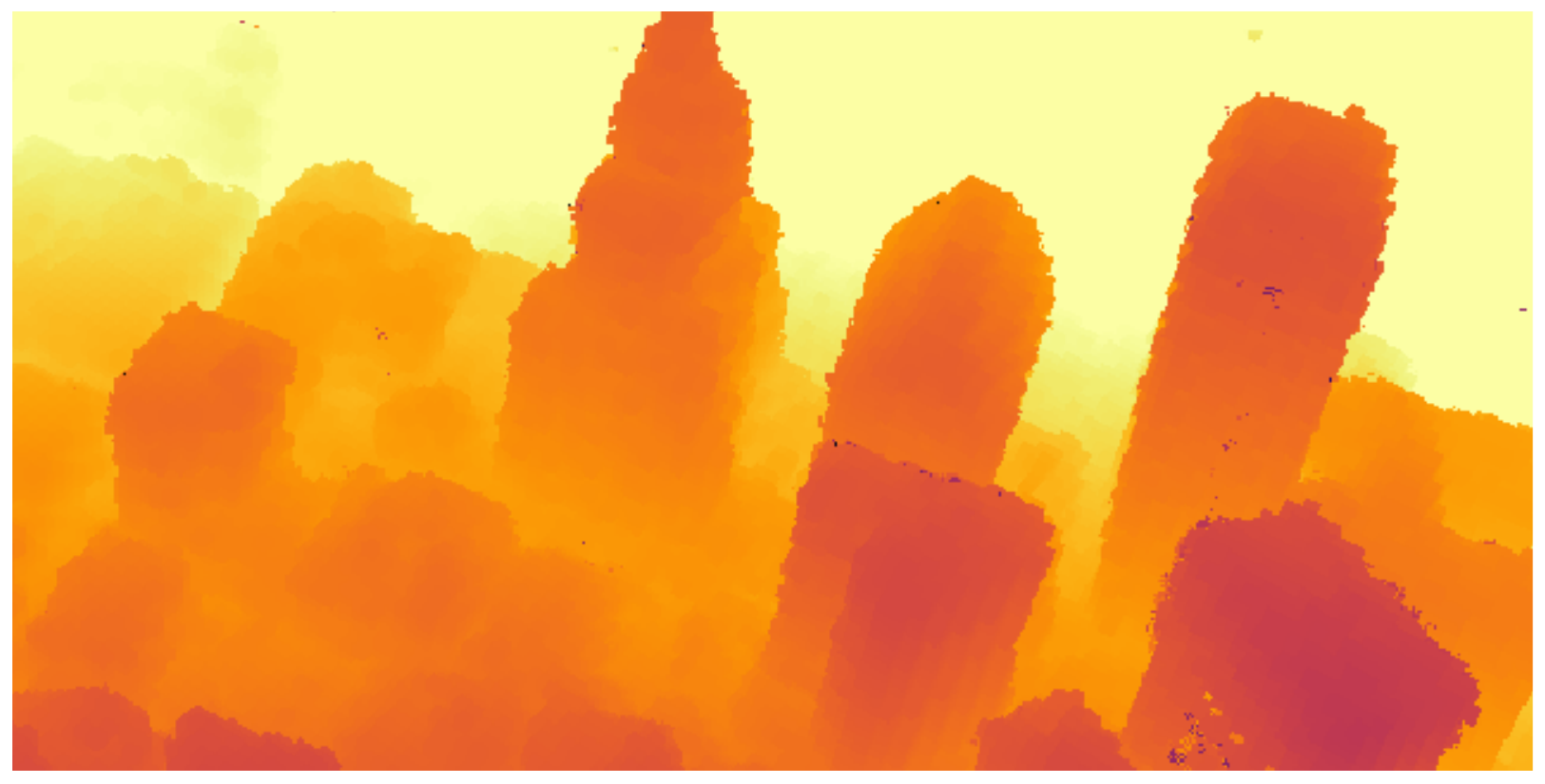}
		\caption[Ours (CNN+MRF)]%
		{{\small Ours (CNN+MRF)}}    
		\label{fig:raynet}
	\end{subfigure}
	\vskip\baselineskip
	\begin{subfigure}[b]{0.3\textwidth}
		\centering
		\includegraphics[width=\textwidth]{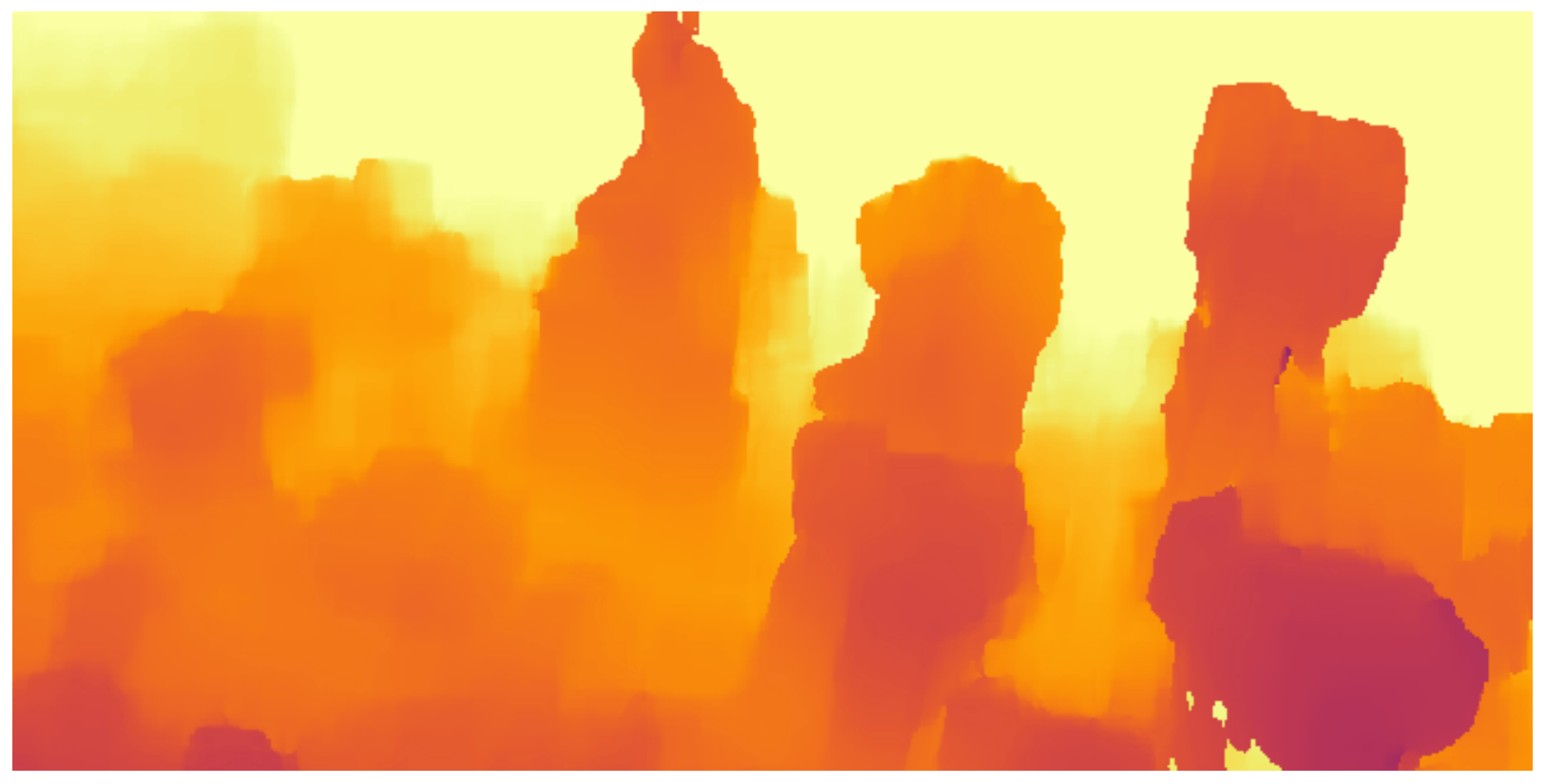}
		\caption[ZNCC]%
		{{\small ZNCC}}
		\label{fig:zncc}
	\end{subfigure}
	\hfill
	\begin{subfigure}[b]{0.3\textwidth}
		\centering
		\includegraphics[width=\textwidth]{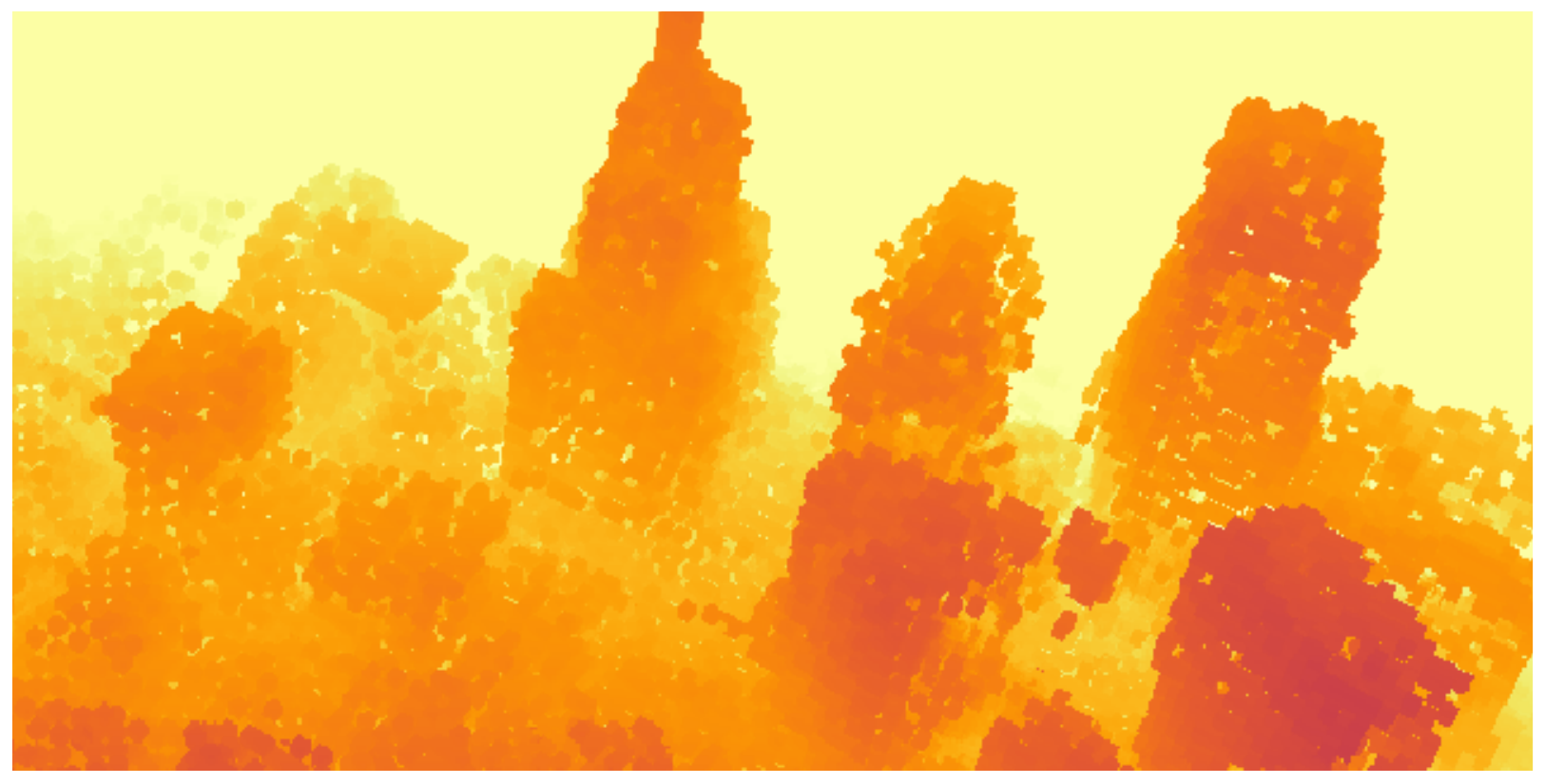}
		\caption[Ulusoy et al. \cite{Ulusoy2015THREEDV}]%
		{{\small Ulusoy et al. \cite{Ulusoy2015THREEDV}}}
		\label{fig:3dv}
	\end{subfigure}
	\hfill
	\begin{subfigure}[b]{0.3\textwidth}
		\centering
		\includegraphics[width=\textwidth]{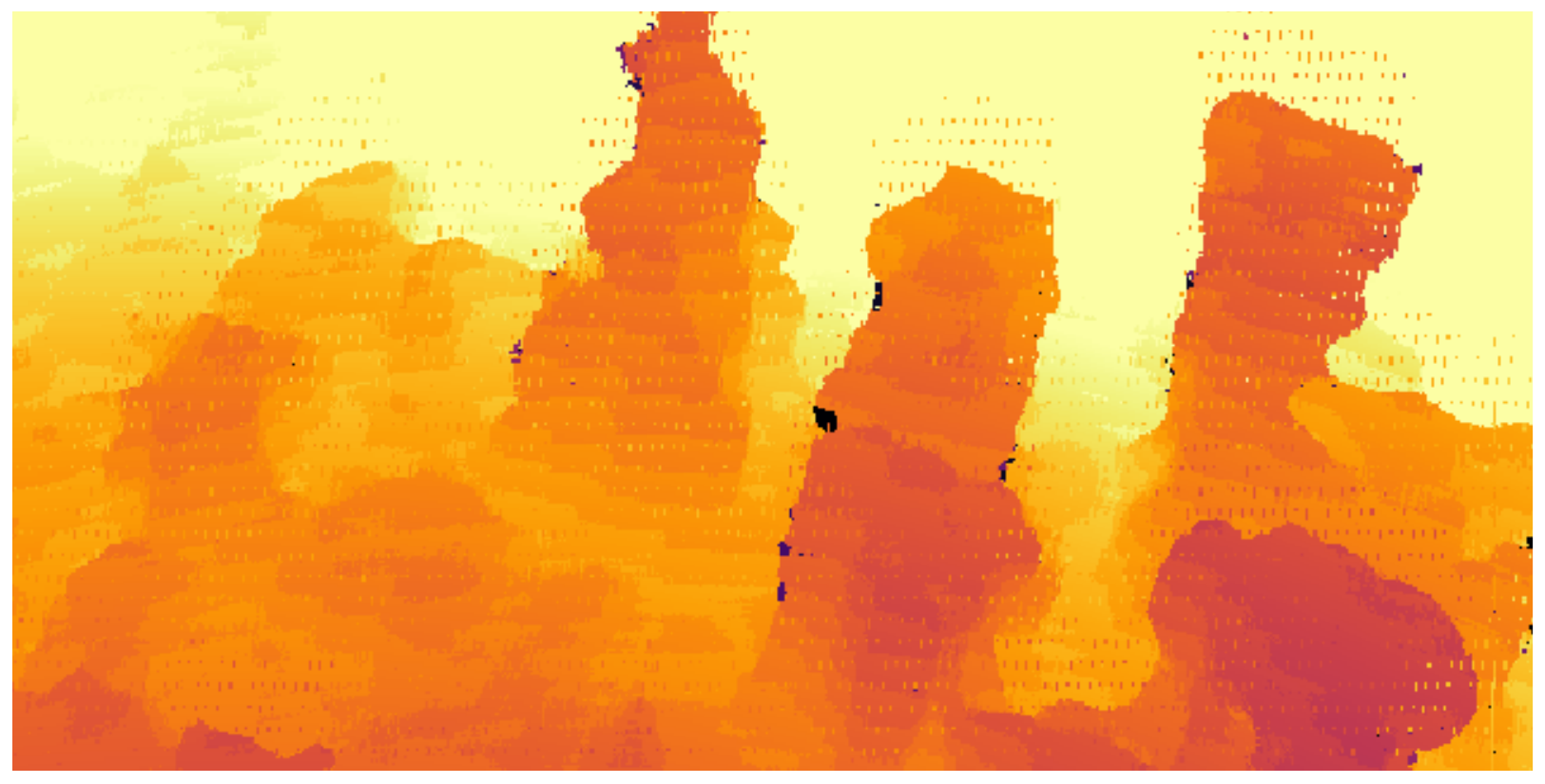}
		\caption[Hartmann et al. \cite{Hartmann2017ICCV}]%
		{{\small Hartmann et al. \cite{Hartmann2017ICCV}}}
		\label{fig:hartmann}
	\end{subfigure}
	\vskip\baselineskip
	\centering
	\begin{subfigure}[b]{0.3\textwidth}
		\centering
		\includegraphics[width=\textwidth]{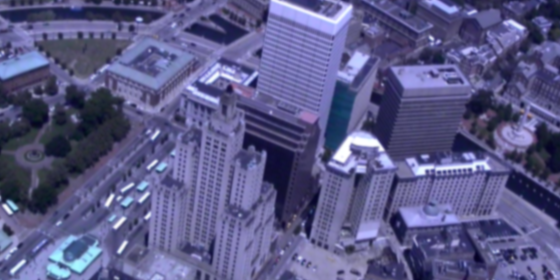}
		\caption[Image]%
		{{\small Image}}
		\label{fig:ground_truth_2}
	\end{subfigure}
	\hfill
	\begin{subfigure}[b]{0.3\textwidth}
		\centering
		\includegraphics[width=\textwidth]{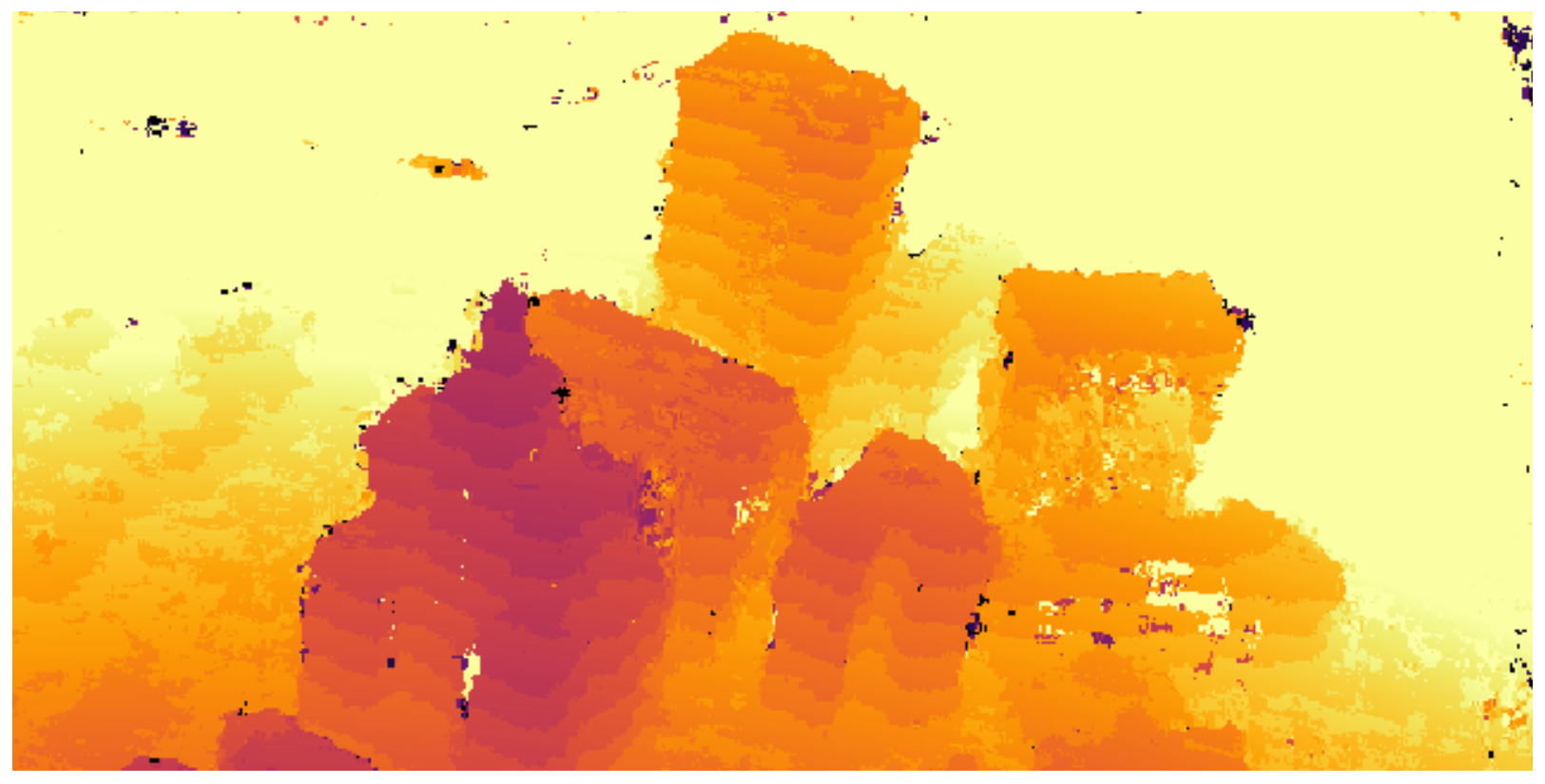}
		\caption[CNN pretrained]%
		{{\small Ours (CNN)}}    
		\label{fig:cnn_pretrained_2}
	\end{subfigure}
	\hfill
	\begin{subfigure}[b]{0.3\textwidth}
		\centering
		\includegraphics[width=\textwidth]{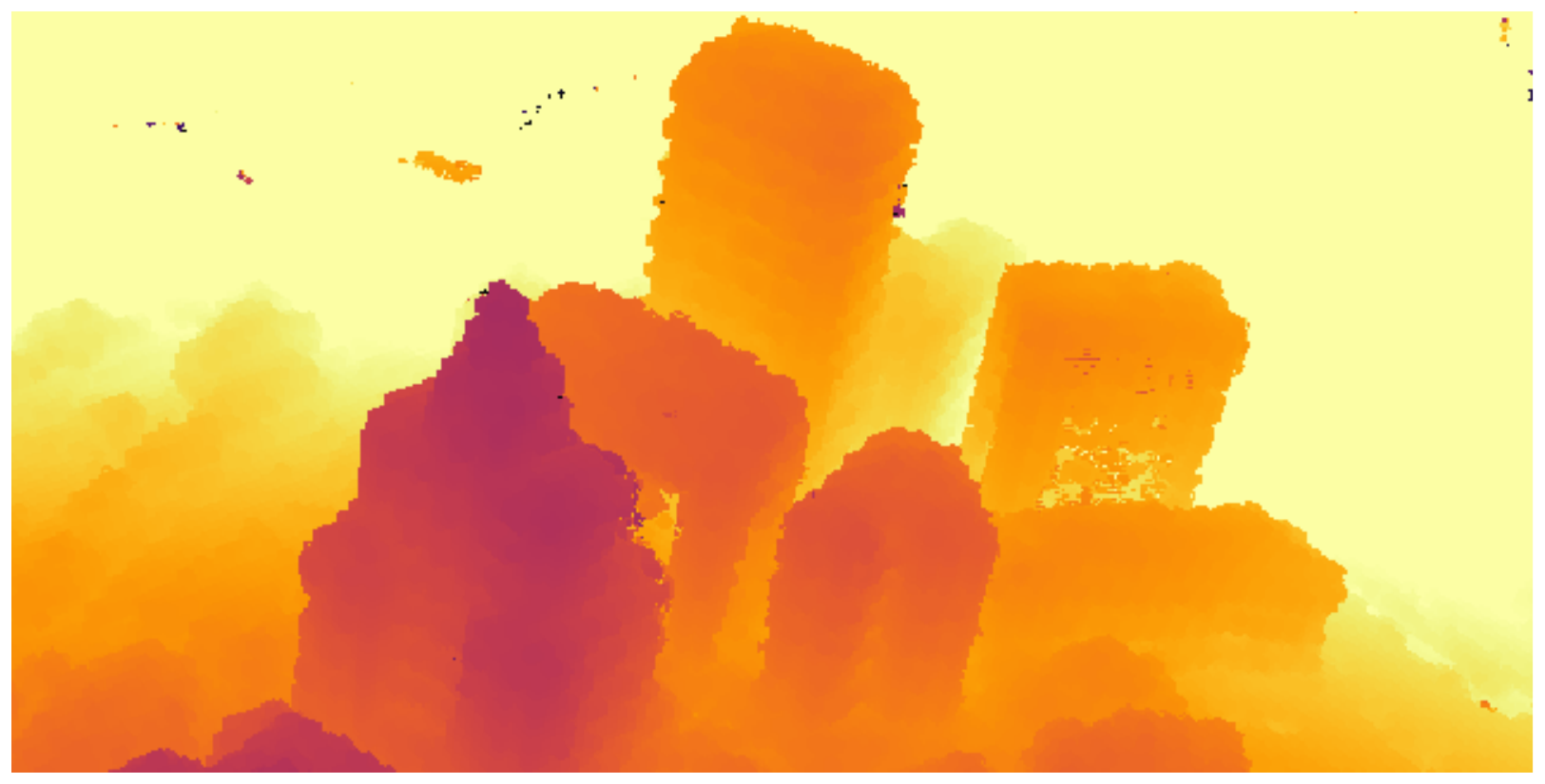}
		\caption[Ours (CNN+MRF)]%
		{{\small Ours (CNN+MRF)}}    
		\label{fig:raynet_2}
	\end{subfigure}
	\vskip\baselineskip
	\begin{subfigure}[b]{0.3\textwidth}
		\centering
		\includegraphics[width=\textwidth]{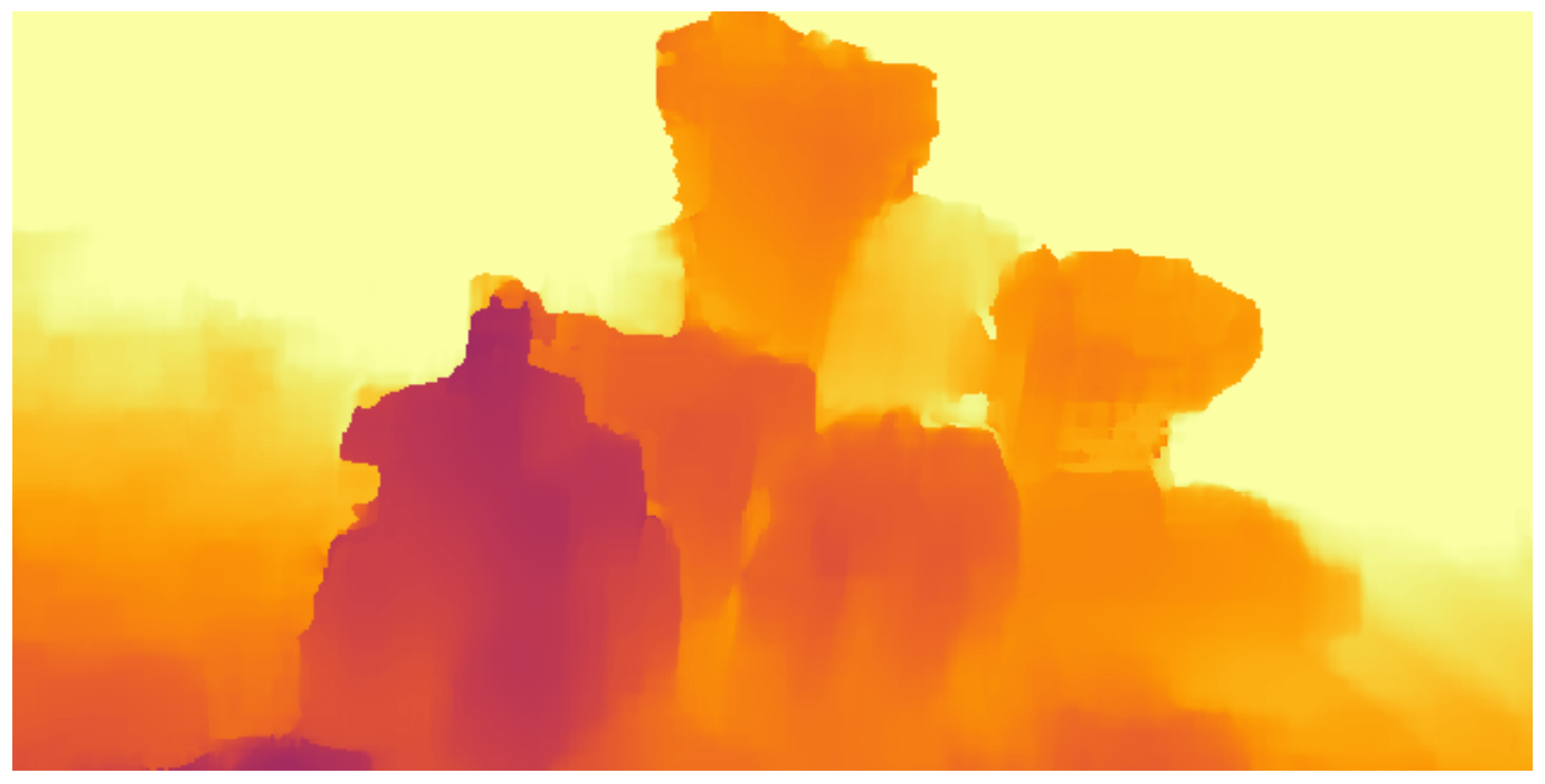}
		\caption[ZNCC]%
		{{\small ZNCC}}
		\label{fig:zncc_2}
	\end{subfigure}
	\hfill
	\begin{subfigure}[b]{0.3\textwidth}
		\centering
		\includegraphics[width=\textwidth]{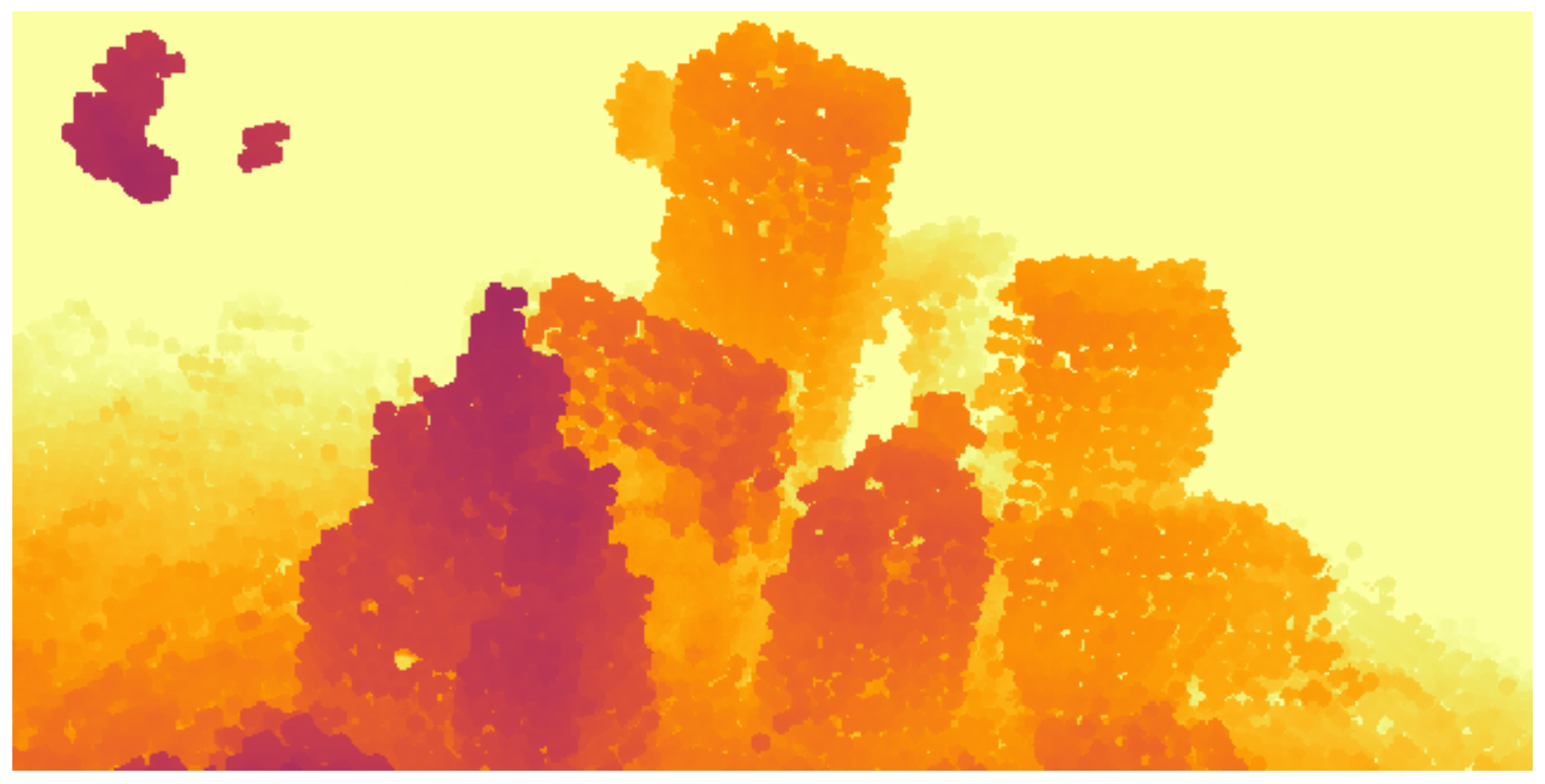}
		\caption[Ulusoy et al. \cite{Ulusoy2015THREEDV}]%
		{{\small Ulusoy et al. \cite{Ulusoy2015THREEDV}}}
		\label{fig:3dv_2}
	\end{subfigure}
	\hfill
	\begin{subfigure}[b]{0.3\textwidth}
		\centering
		\includegraphics[width=\textwidth]{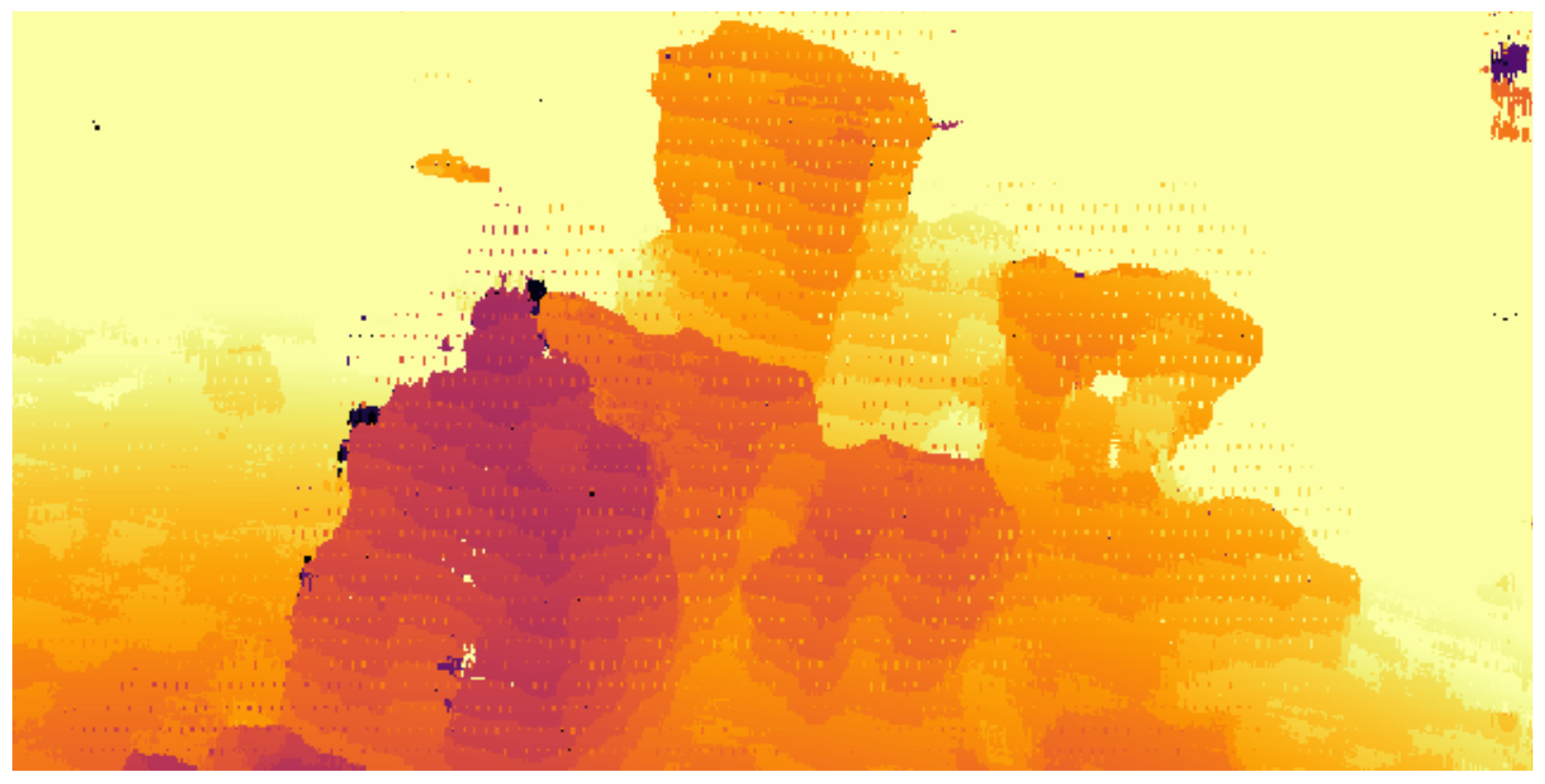}
		\caption[Hartmann et al. \cite{Hartmann2017ICCV}]%
		{{\small Hartmann et al. \cite{Hartmann2017ICCV}}}
		\label{fig:hartmann_2}
	\end{subfigure}
	\caption{{\bf Qualitative Results on Aerial Dataset}. We show the depth maps predicted by our method (b+c,i+h) as well as three baselines (d-f,j-l) for two input images from different viewpoints (a,g). Darker colors correspond to closer regions.}
	\label{fig:qualitative_restrepo}
\end{figure*}

In this Section, we validate our technique on the aerial dataset of Restrepo \etal \cite{Restrepo2014JPRS} and compare it to various model-based and learning-based baselines.

\boldparagraph{Quantitative evaluation}
We pretrain the neural network using the Adam optimizer \cite{Kingma2015ICLR} with a learning rate of $10^{-3}$ and a batch size of 32 for 100K iterations. Subsequently, for end-to-end training, we use a voxel grid of size $64^3$, the same optimizer with learning rate $10^{-4}$ and batches of $2000$ randomly sampled rays from consecutive images. RayNet is trained for $20$K iterations. Note that due to the use of the larger batch size, RayNet is trained on approximately 10 times more rays compared to pretraining. During training we use a voxel grid of size $64^3$ to reduce computation. However, we run the forward pass with a voxel grid of size $256^3$ for the final evaluation.

We compare RayNet with two commonly used patch comparison measures: the sum of absolute differences (SAD) and the zero-mean normalized cross correlation (ZNCC). In particular, we use SAD or ZNCC to compute a cost volume per pixel and then choose the depth with the lowest cost to produce a depthmap. We use the publicly available code distributed by Haene~\etal \cite{Haene2014THREEDV} for these two baselines. We follow \cite{Haene2014THREEDV} and compute the max score among all pairs of patch combinations, which allows some robustness to occlusions. 
In addition, we also compare our method with the probabilistic  3D reconstruction method by Ulusoy et al. \cite{Ulusoy2015THREEDV}, again using their publicly available implementation.
We further compare against the learning based approach of
Hartmann \etal~\cite{Hartmann2017ICCV}, which we reimplemented and trained based on the original paper \cite{Hartmann2017ICCV}.

\begin{table*}
    \centering
    \begin{tabular}{|l|C{1.4cm}|C{1.4cm}|C{1.4cm}|C{1.4cm}|C{1.4cm}|C{1.4cm}|C{1.4cm}|}
        \hline
        Methods & \multicolumn{2}{c|}{Accuracy} &
          \multicolumn{2}{c|}{Completeness} &
          \multicolumn{2}{c|}{Mean Depth Error} &
          Chamfer \\
        \hline
        & Mean & Median & Mean & Median & Mean & Median & \\
        \hline
        ZNCC & 0.0753 & 0.0384 & 0.0111 & \textbf{0.0041} & 0.1345 & 0.1351 & 0.0432\\
        \hline
        SAD & 0.0746 & 0.0373 & 0.0120 & 0.0042 & 0.1233 & 0.1235 & 0.0433\\
        \hline
        Ulusoy et al. \cite{Ulusoy2015THREEDV} & 0.0790 & 0.0167 & \textbf{0.0088} & 0.0065 & 
            0.1143 & 0.1050 & 0.0439\\
        \hline
        Hartmann et al. \cite{Hartmann2017ICCV} & 0.0907 & 0.0285 & 0.0209 & 0.0209 & 0.1648 &  0.1222 & 0.0558\\
        \hline
        Ours (CNN) & 0.0804 & 0.0220 & 0.0154 & 0.0132 & 0.0977 & 0.0916 & 0.0479\\
        \hline
        Ours (CNN+MRF) & \textbf{0.0611} & \textbf{0.0160} & 0.0125 & 0.0085 & \textbf{0.0744} & \textbf{0.0728} & \textbf{0.0368}\\
        \hline
    \end{tabular}
    \caption{{\bf Quantitative Results on Aerial Dataset.} We show the mean and median accuracy, completeness, per pixel error and Chamfer distance for various baselines and our method. Lower values indicate better results. See text for details.
    }
    \label{tab:experimental_setup_1}
    \vskip -10pt
\end{table*}

\tabref{tab:experimental_setup_1} summarizes accuracy, completeness and per pixel mean depth error for all implemented baselines. In addition to the aforementioned baselines, we also compare our full RayNet approach {\it Ours (CNN+MRF)} with our CNN frontend in isolation, denoted {\it Ours (CNN)}.
We observe that joint optimization of the full model improves upon the CNN frontend.
Furthermore, RayNet outperforms both the classic as well as learning-based baselines in terms of accuracy, mean depth error and Chamfer distance while performing on par with most baselines in terms of completeness.

\boldparagraph{Qualitative Evaluation}
We visualize the results of RayNet and the baselines in  Fig.~\ref{fig:qualitative_restrepo}.
Both the ZNCC baseline and the approach of Hartmann \etal \cite{Hartmann2017ICCV} require a large receptive field size for optimal performance, yielding smooth depth maps, However, this large receptive field also causes bleeding artefacts at object boundaries. In contrast, our baseline CNN and the approach of Ulusoy \etal \cite{Ulusoy2015THREEDV} yield sharper boundaries, while exhibiting a larger level of noise.
By combining the advantages of learning-based descriptors with a small receptive field and MRF inference, our full RayNet approach (CNN+MRF) results in significantly smoother reconstructions while retaining sharp object boundaries.
Additional results are provided in the supplementary material.

\subsection{DTU Dataset}

In this section, we provide results on the  BIRD and BUDDHA scenes of the DTU dataset. We use the provided splits to train and test RayNet. We evaluate SurfaceNet~\cite{Ji2017ICCV} at two different resolutions: the original high resolution variant, which requires more than $4$  hours to reconstruct a single scene, and a faster variant that uses approximately the same resolution ($256^3$) as our approach. We refer to the first one as \emph{SurfaceNet (HD)} and to the latter as \emph{SurfaceNet (LR)}. We test SurfaceNet with the pretrained models provided by \cite{Ji2017ICCV}.

We evaluate both methods in terms of accuracy and completeness and report the mean and the median in \tabref{tab:experimental_setup_2}. Our full RayNet model outperforms nearly all baselines in terms of completeness, while it performs worse in terms of accuracy.
We believe this is due to the fact that both \cite{Ji2017ICCV} and \cite{Hartmann2017ICCV} utilize the original high resolution images, while our approach operates on downsampled versions of the images. We observe that some of the fine textural details that are present in the original images are lost in the down-sampled versions. Besides, our current implementation of the 2D feature extractor uses a smaller receptive field size ($11 \time 11$ pixels) compared to SurfaceNet ($64\times64$) and \cite{Hartmann2017ICCV} ($32\times32$). Finally, while our method aims at predicting a complete 3D reconstruction inside the evaluation volume (resulting in occasional outliers), \cite{Ji2017ICCV} and \cite{Hartmann2017ICCV} prune unreliable predictions from the output and return an incomplete reconstruction (resulting in higher accuracy and lower completeness).
 
\begin{figure}[t!]
	\centering
	\begin{subfigure}[b]{0.30\linewidth}
    	\centering
        \includegraphics[width=\textwidth]{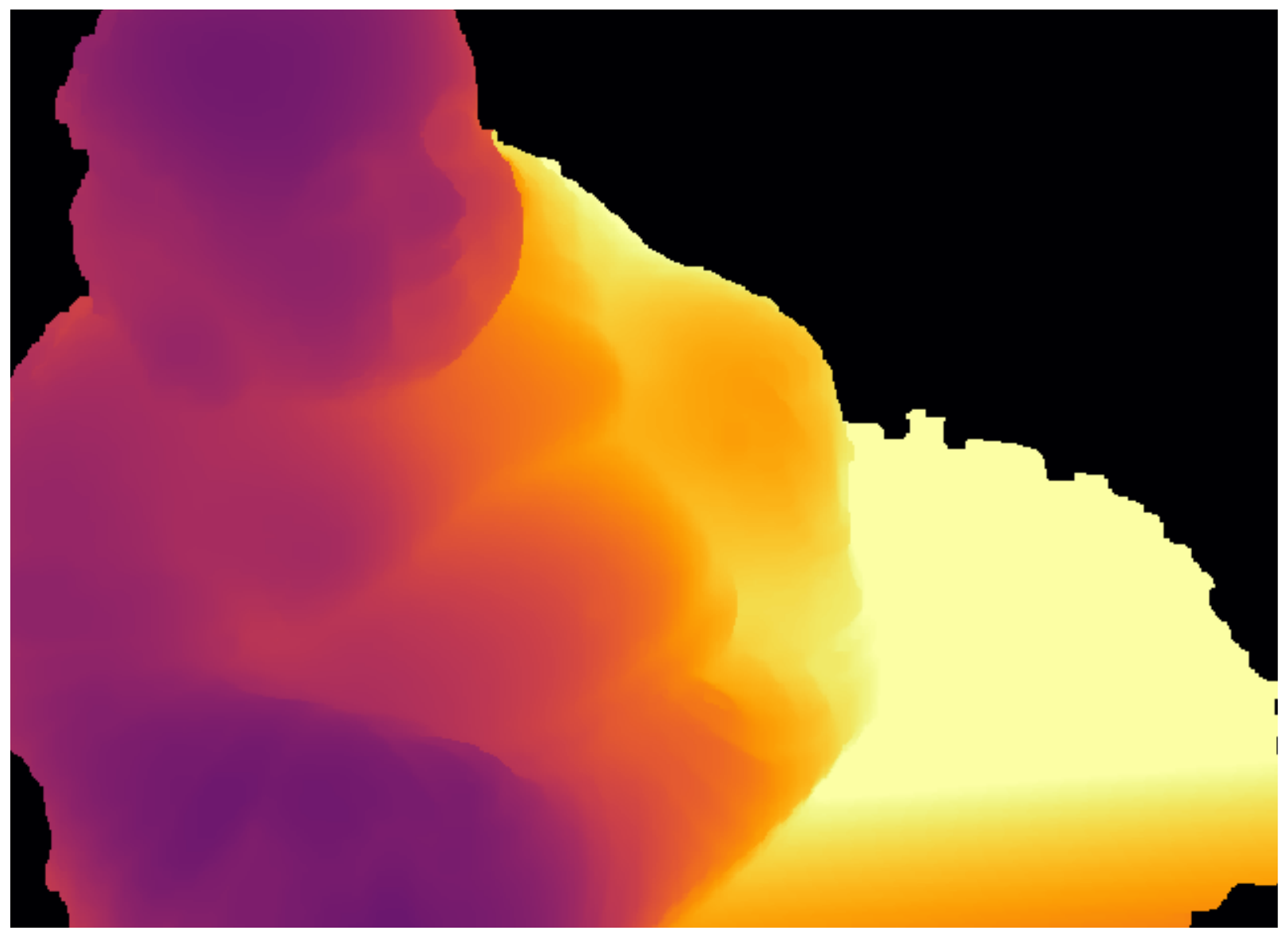}
        \caption[Ground Truth]%
        {{\small Ground Truth}}
    \end{subfigure}
    \hfill
    \begin{subfigure}[b]{0.30\linewidth}
    	\centering
        \includegraphics[width=\textwidth]{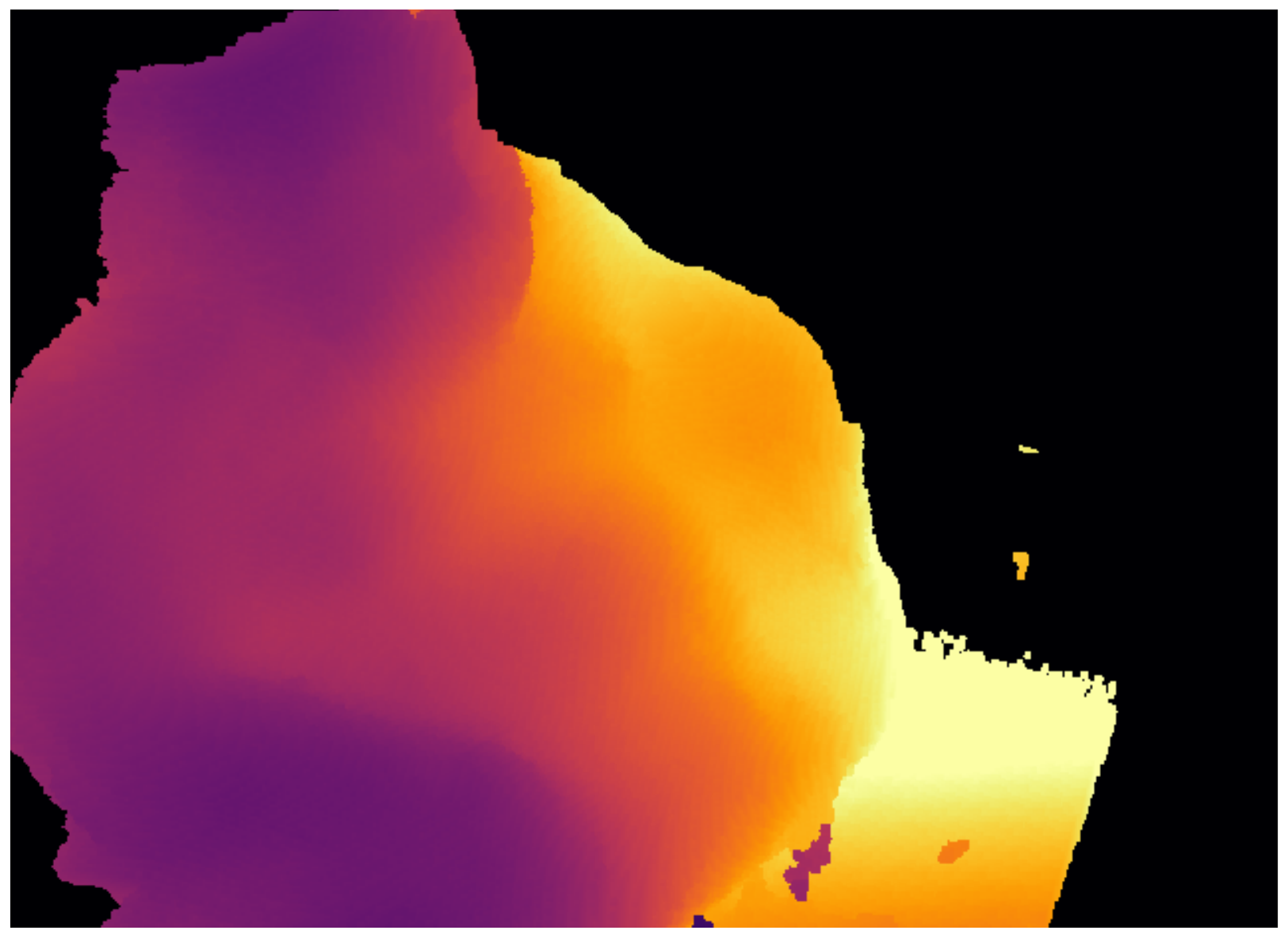}
        \caption[SurfaceNet \cite{Ji2017ICCV}]%
        {{\small SurfaceNet \cite{Ji2017ICCV}}}    
    \end{subfigure}
    \hfill
    \begin{subfigure}[b]{0.30\linewidth}
    	\centering
        \includegraphics[width=\textwidth]{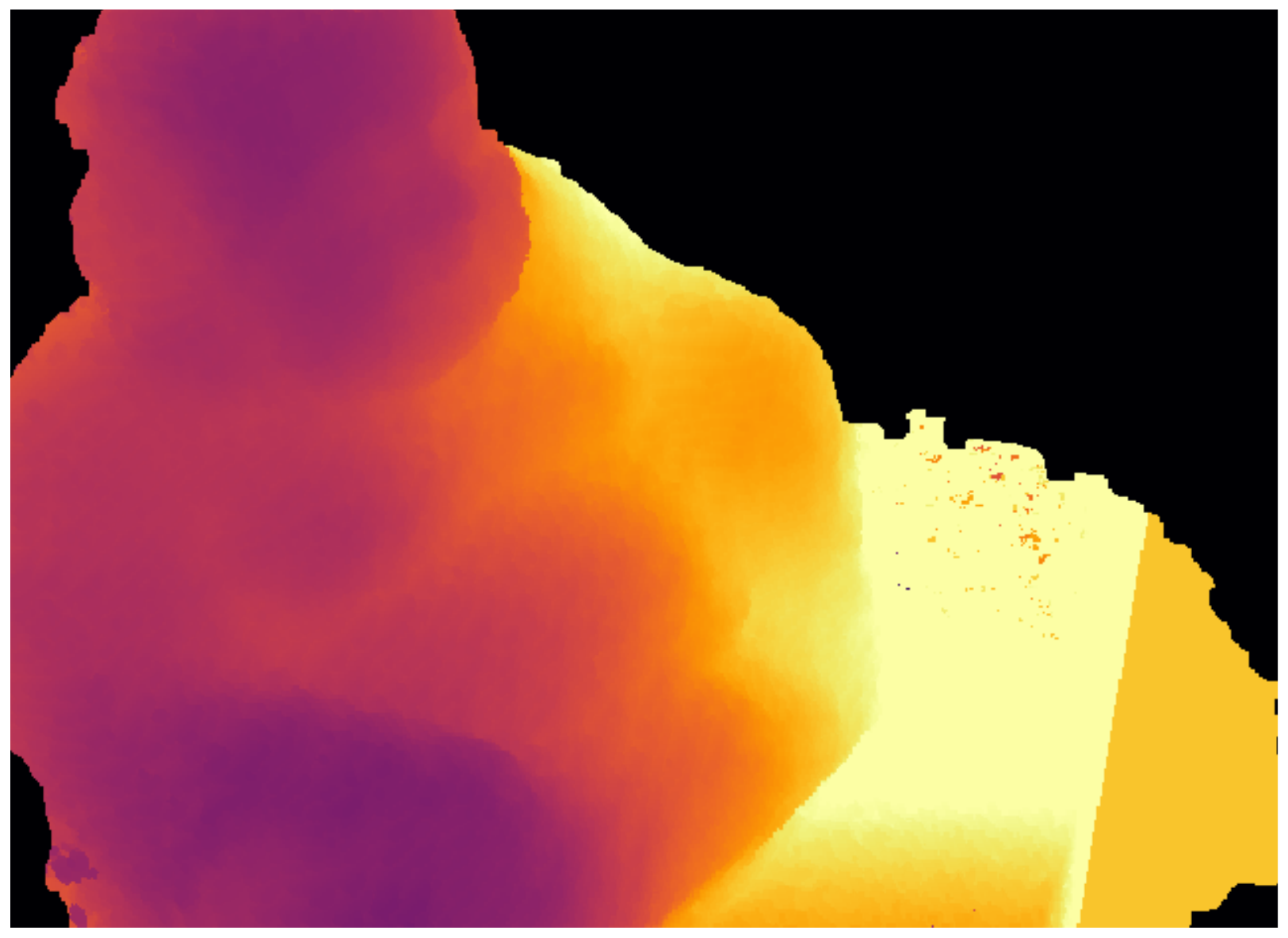}
        \caption[Ours]%
        {{\small RayNet}}    
    \end{subfigure}
	\begin{subfigure}[b]{0.30\linewidth}
    	\centering
        \includegraphics[width=\textwidth]{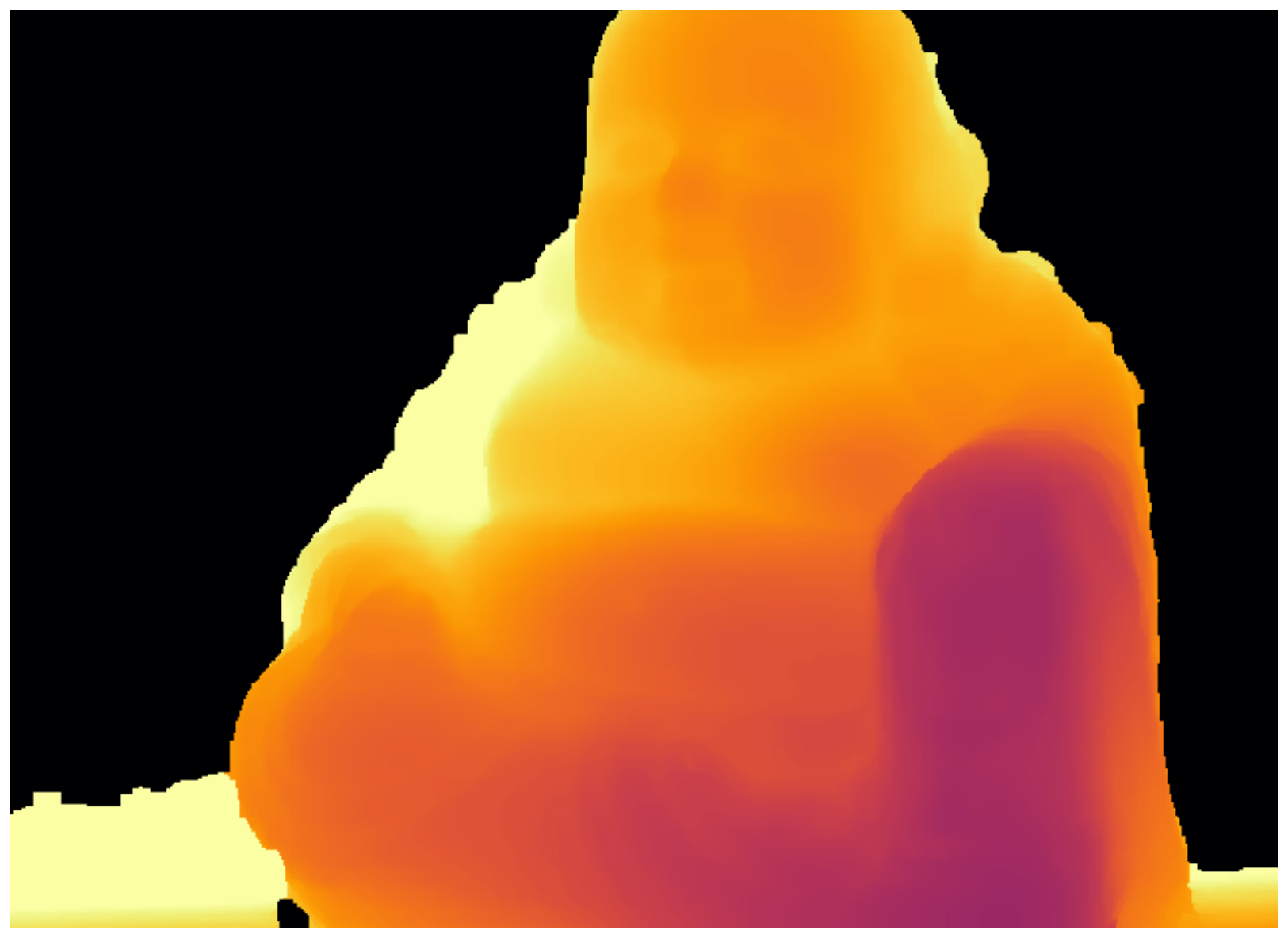}
        \caption[Ground Truth]%
        {{\small Ground Truth}}
    \end{subfigure}
    \hfill
    \begin{subfigure}[b]{0.30\linewidth}
    	\centering
        \includegraphics[width=\textwidth]{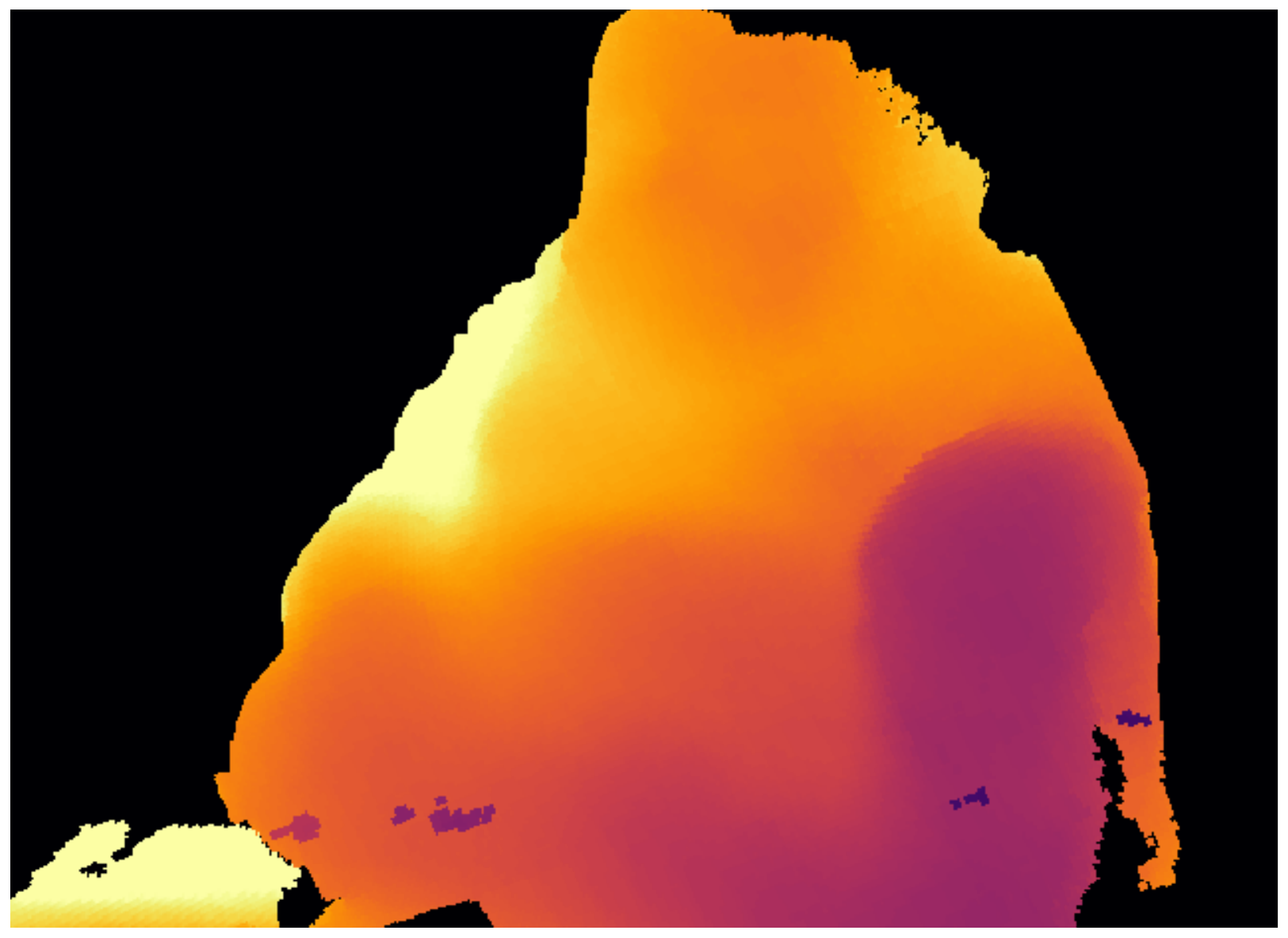}
        \caption[SurfaceNet \cite{Ji2017ICCV}]%
        {{\small SurfaceNet \cite{Ji2017ICCV}}}    
    \end{subfigure}
    \hfill
    \begin{subfigure}[b]{0.30\linewidth}
    	\centering
        \includegraphics[width=\textwidth]{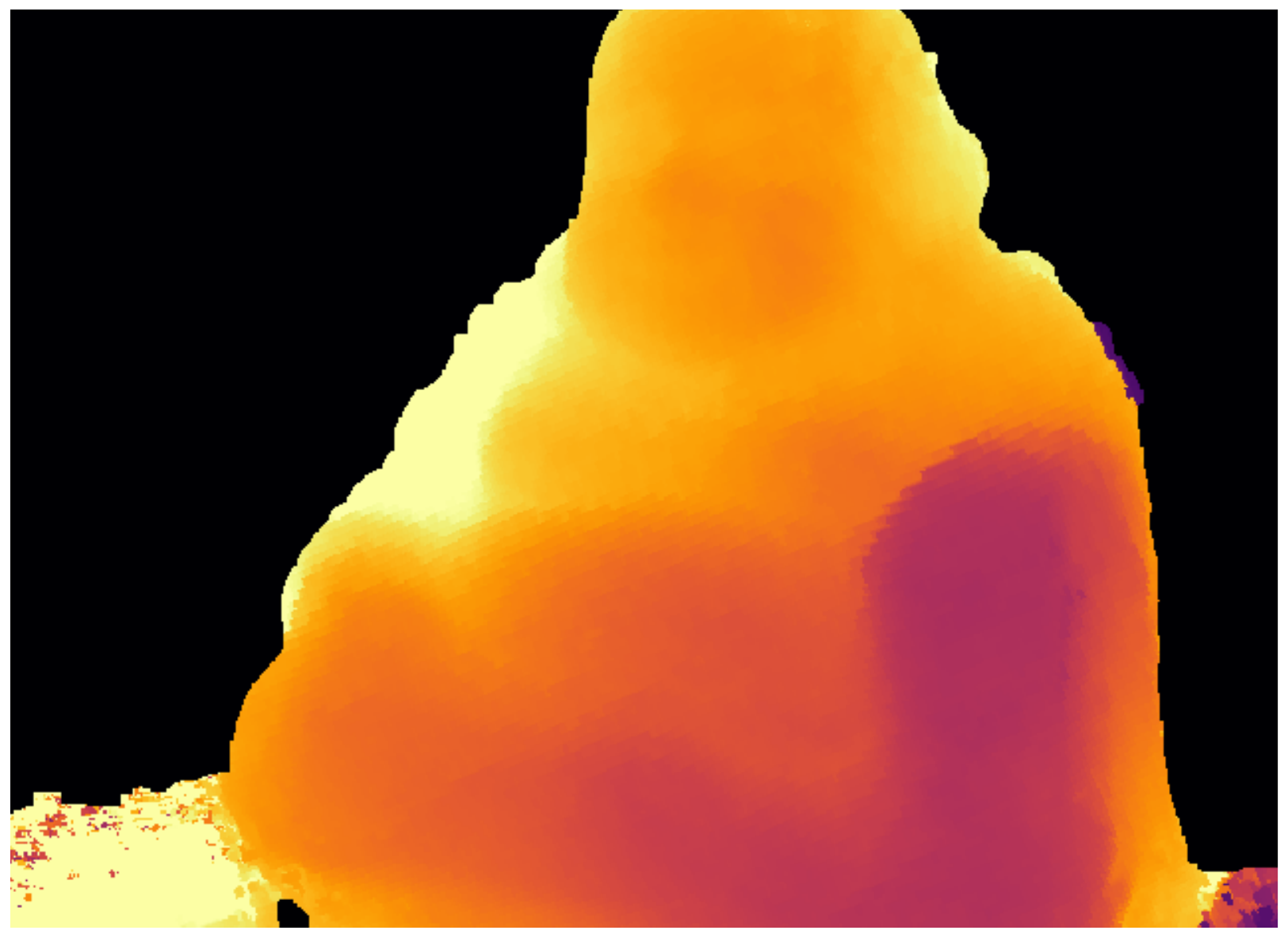}
        \caption[RayNet]%
        {{\small RayNet}}    
    \end{subfigure}
	\begin{subfigure}[b]{0.30\linewidth}
    	\centering
        \includegraphics[width=\textwidth]{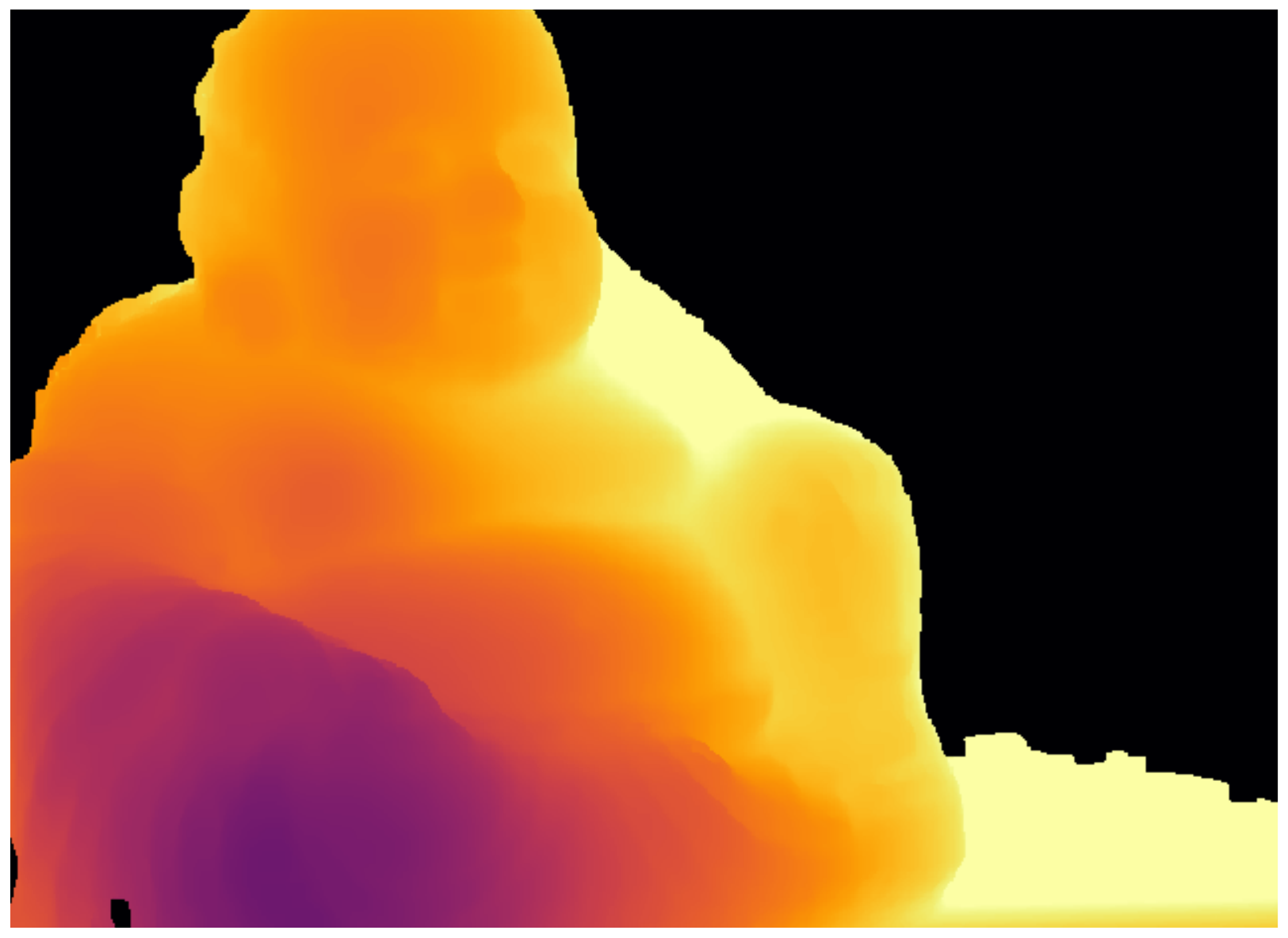}
        \caption[Ground Truth]%
        {{\small Ground Truth}}
    \end{subfigure}
    \hfill
    \begin{subfigure}[b]{0.30\linewidth}
    	\centering
        \includegraphics[width=\textwidth]{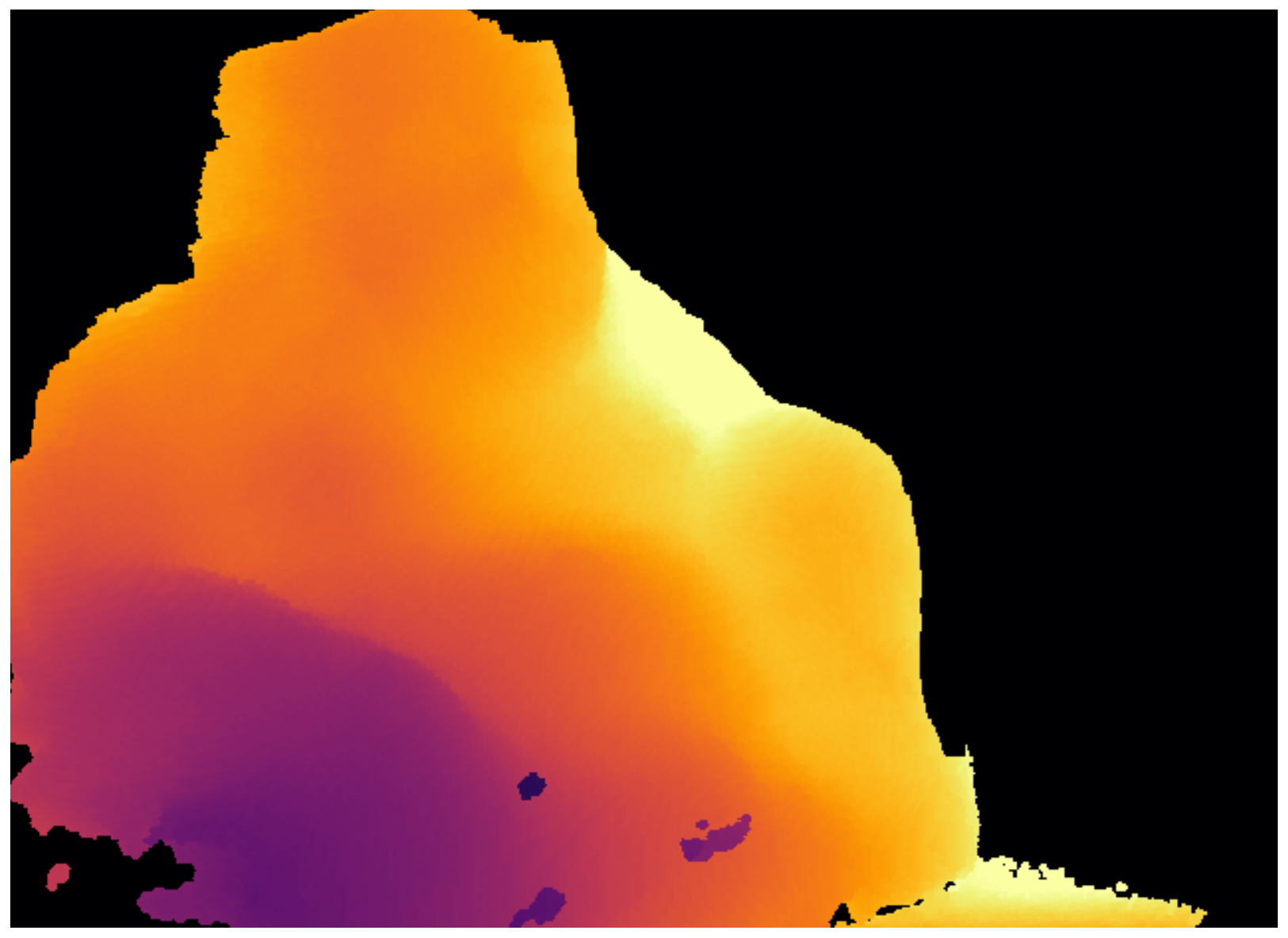}
        \caption[SurfaceNet \cite{Ji2017ICCV}]%
        {{\small SurfaceNet \cite{Ji2017ICCV}}}    
    \end{subfigure}
    \hfill
    \begin{subfigure}[b]{0.30\linewidth}
    	\centering
        \includegraphics[width=\textwidth]{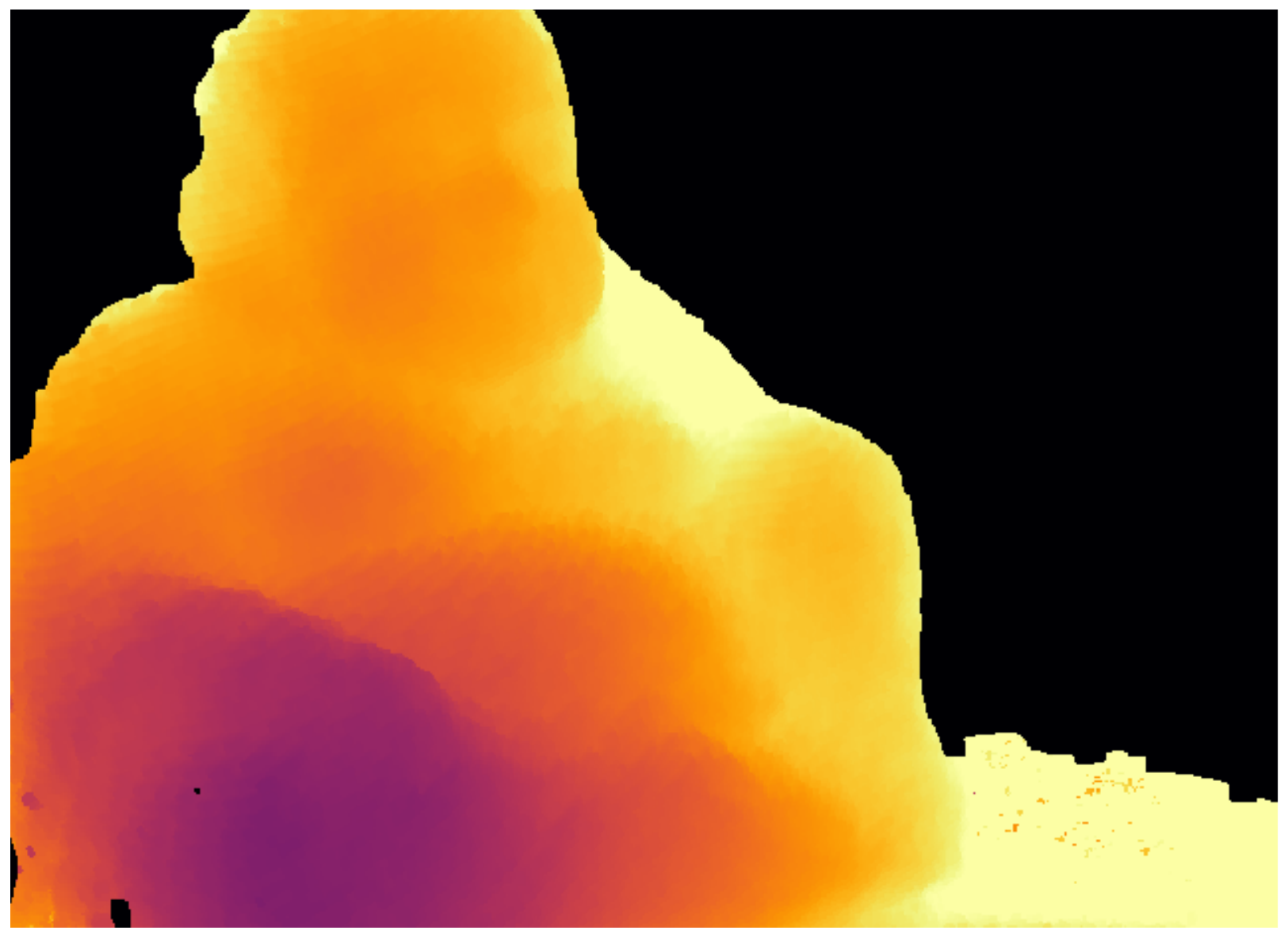}
        \caption[RayNet]%
        {{\small RayNet}}    
    \end{subfigure}

	\begin{subfigure}[b]{0.30\linewidth}
    	\centering
        \includegraphics[width=\textwidth]{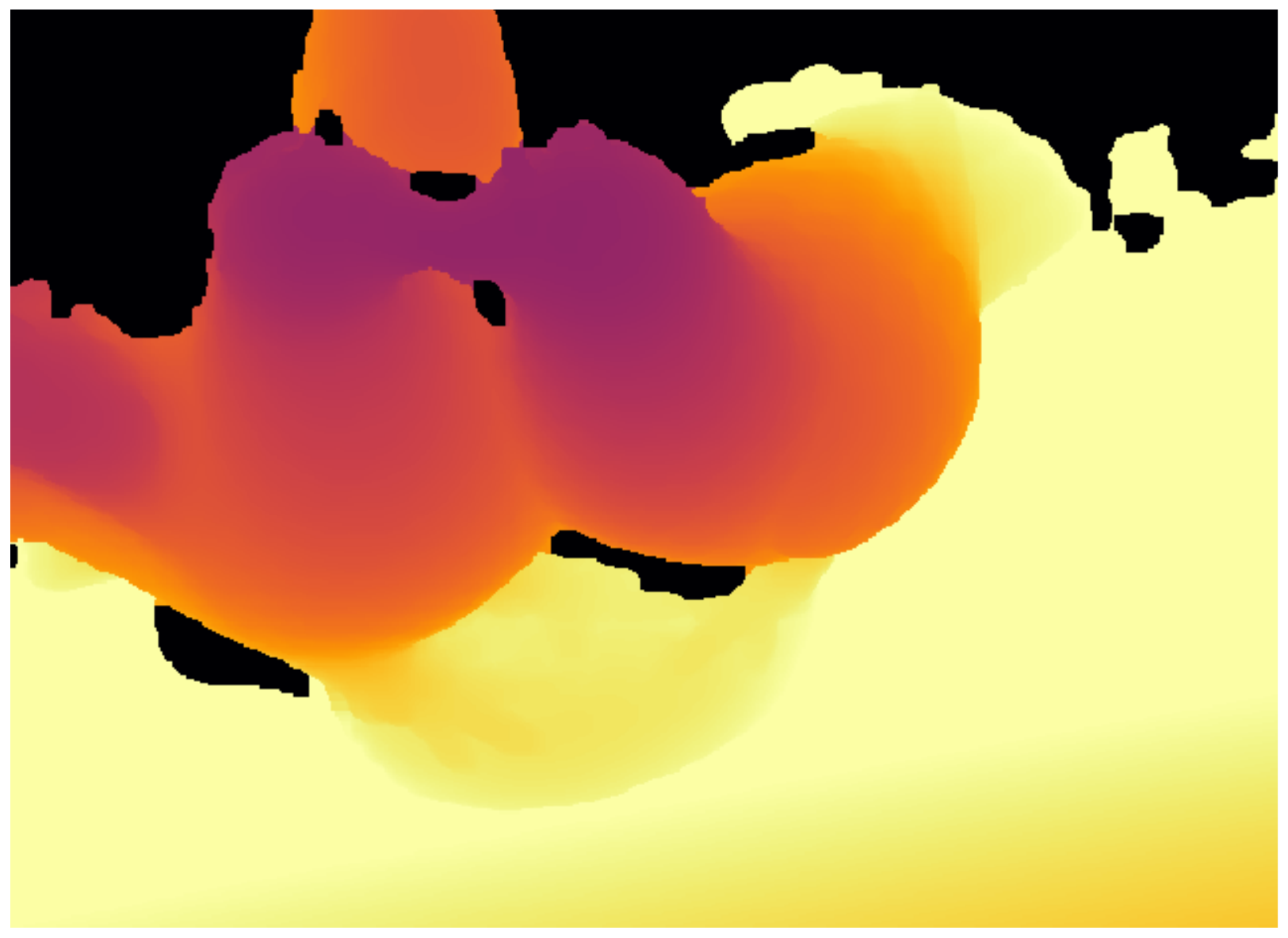}
        \caption[Ground Truth]%
        {{\small Ground Truth}}
    \end{subfigure}
    \hfill
    \begin{subfigure}[b]{0.30\linewidth}
    	\centering
        \includegraphics[width=\textwidth]{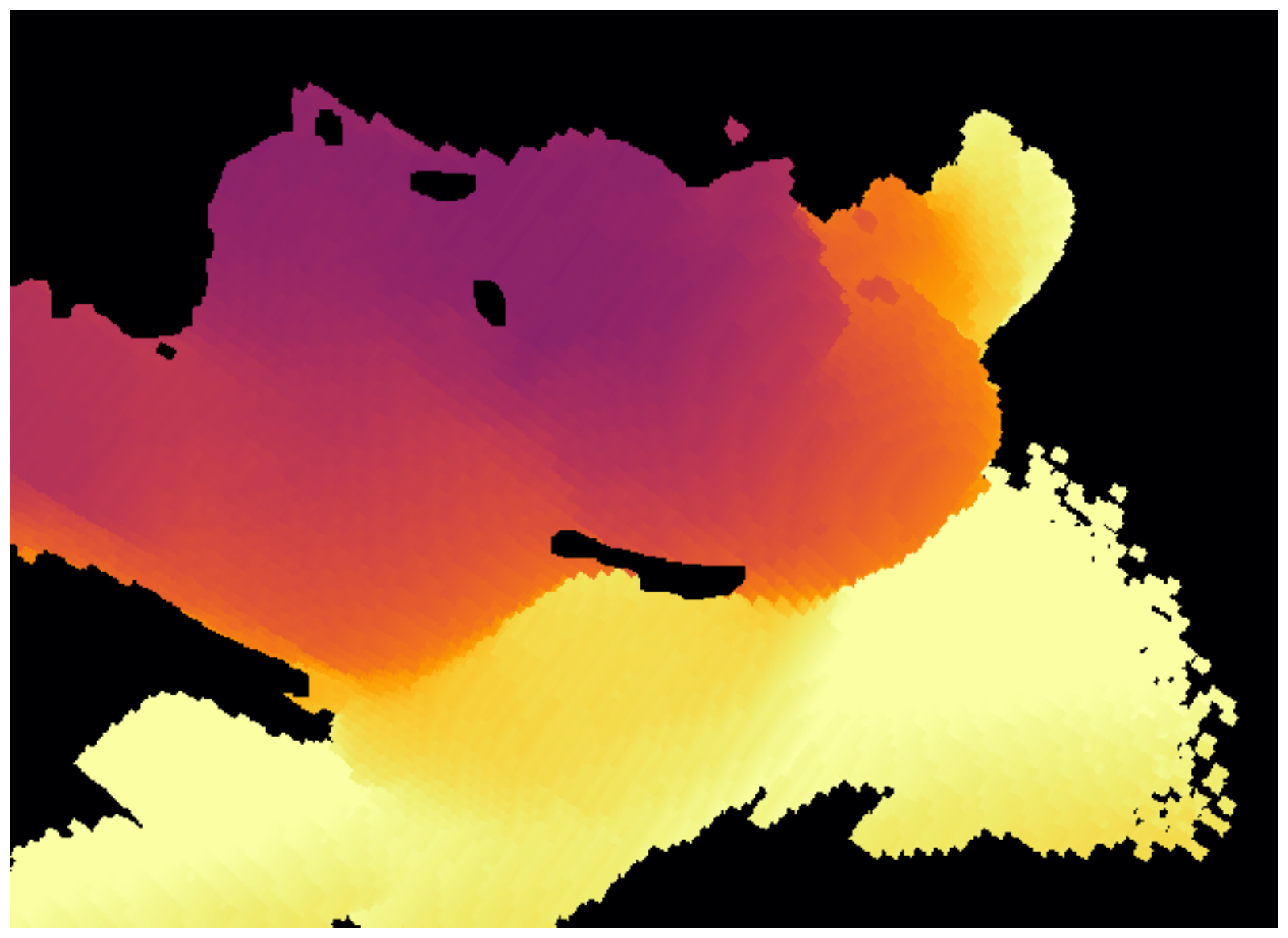}
        \caption[SurfaceNet \cite{Ji2017ICCV}]%
        {{\small SurfaceNet \cite{Ji2017ICCV}}}    
    \end{subfigure}
    \hfill
    \begin{subfigure}[b]{0.30\linewidth}
    	\centering
        \includegraphics[width=\textwidth]{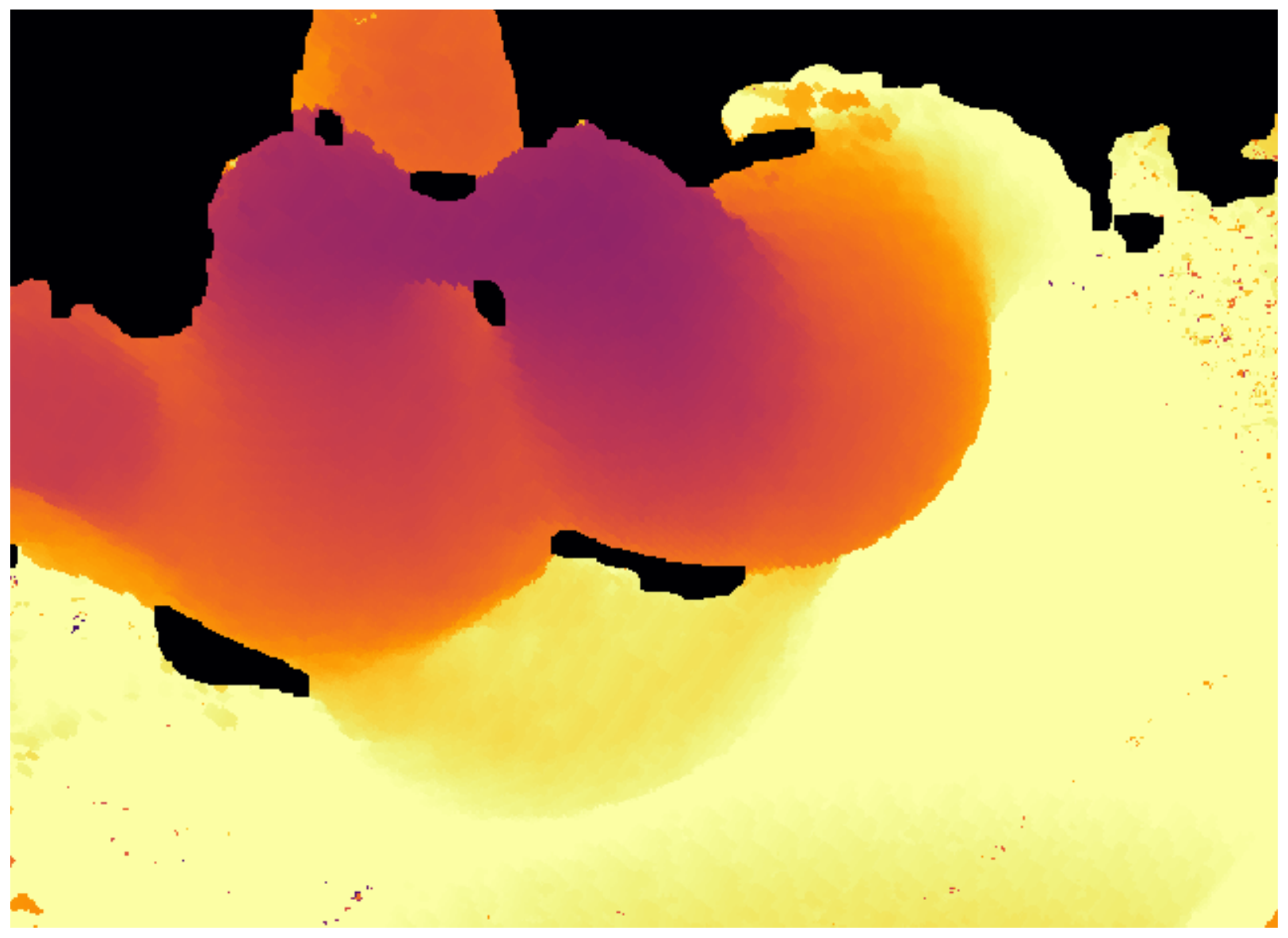}
        \caption[RayNet]%
        {{\small RayNet}}    
    \end{subfigure}
	\begin{subfigure}[b]{0.30\linewidth}
    	\centering
        \includegraphics[width=\textwidth]{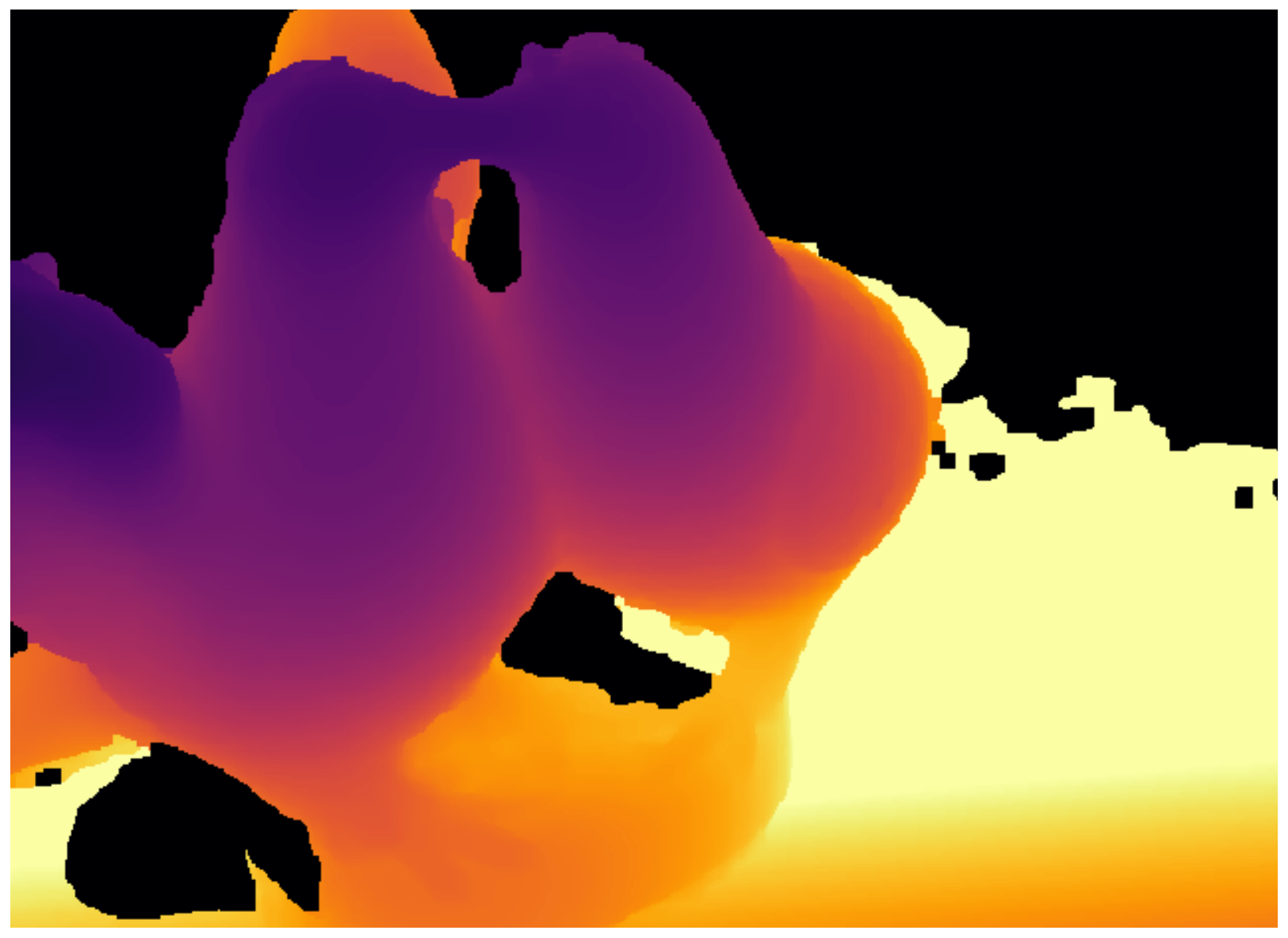}
        \caption[Ground Truth]%
        {{\small Ground Truth}}
    \end{subfigure}
    \hfill
    \begin{subfigure}[b]{0.30\linewidth}
    	\centering
        \includegraphics[width=\textwidth]{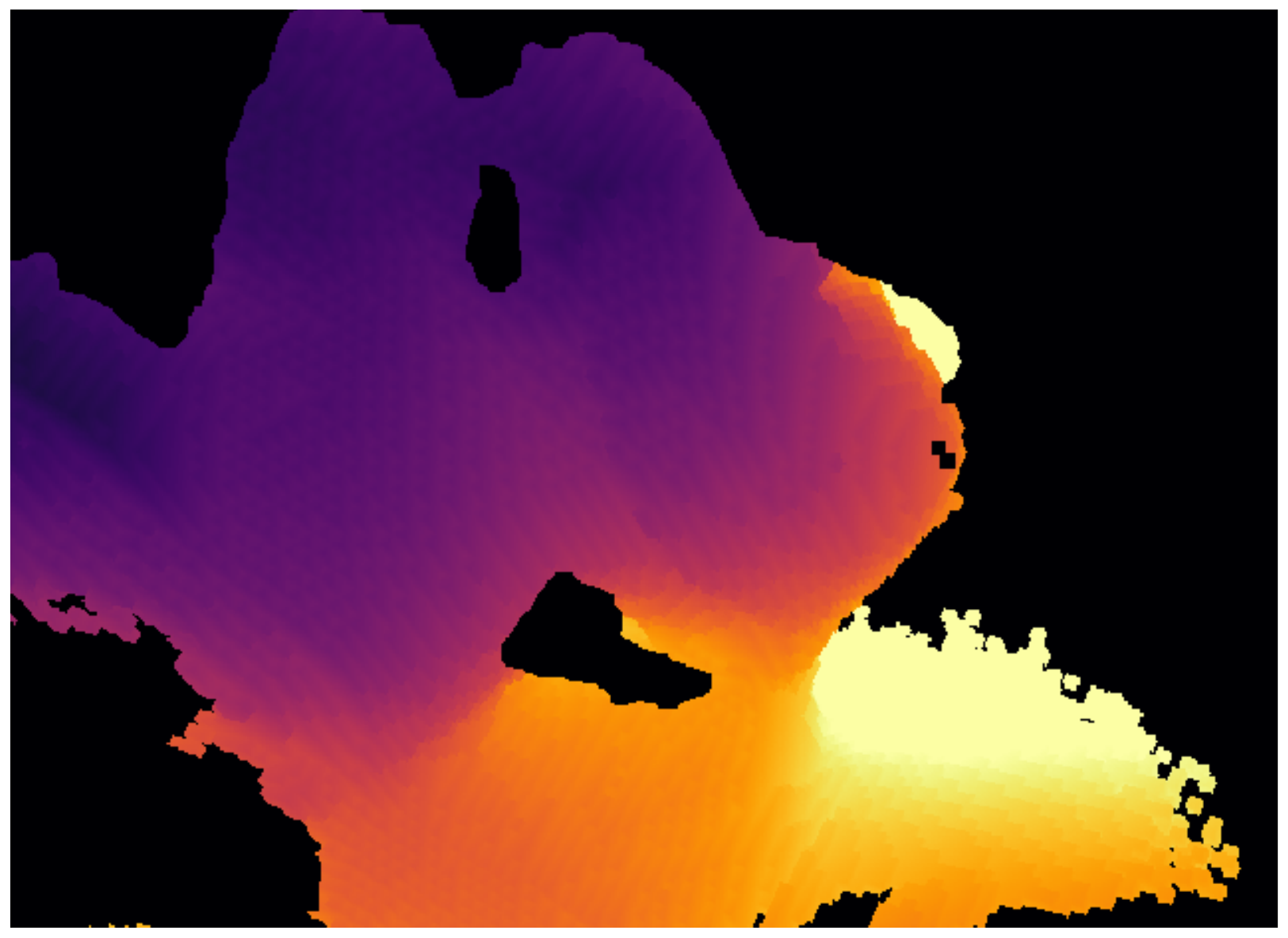}
        \caption[SurfaceNet \cite{Ji2017ICCV}]%
        {{\small SurfaceNet \cite{Ji2017ICCV}}}    
    \end{subfigure}
    \hfill
    \begin{subfigure}[b]{0.30\linewidth}
    	\centering
        \includegraphics[width=\textwidth]{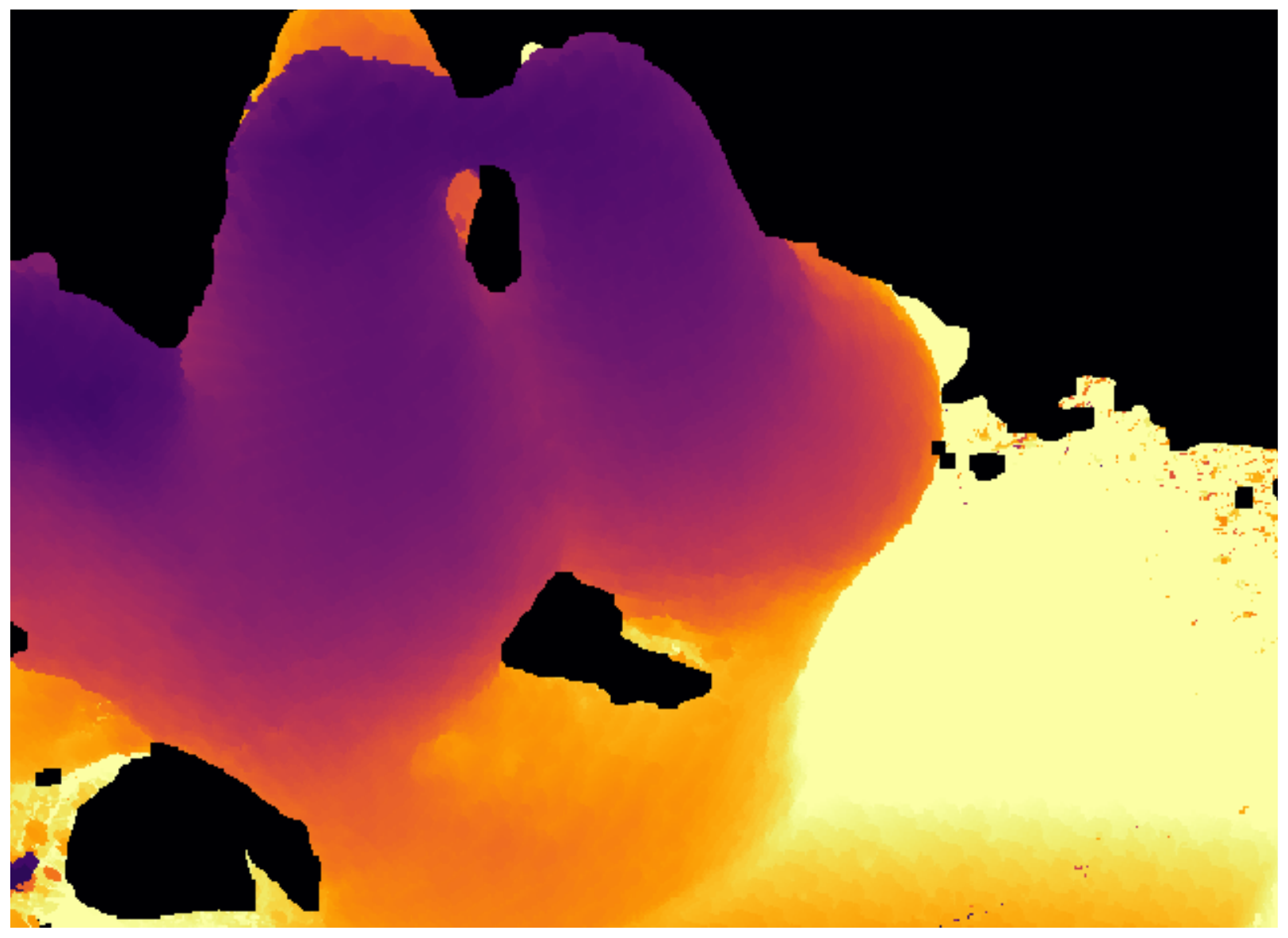}
        \caption[RayNet]%
        {{\small RayNet}}    
    \end{subfigure}
	\begin{subfigure}[b]{0.30\linewidth}
    	\centering
        \includegraphics[width=\textwidth]{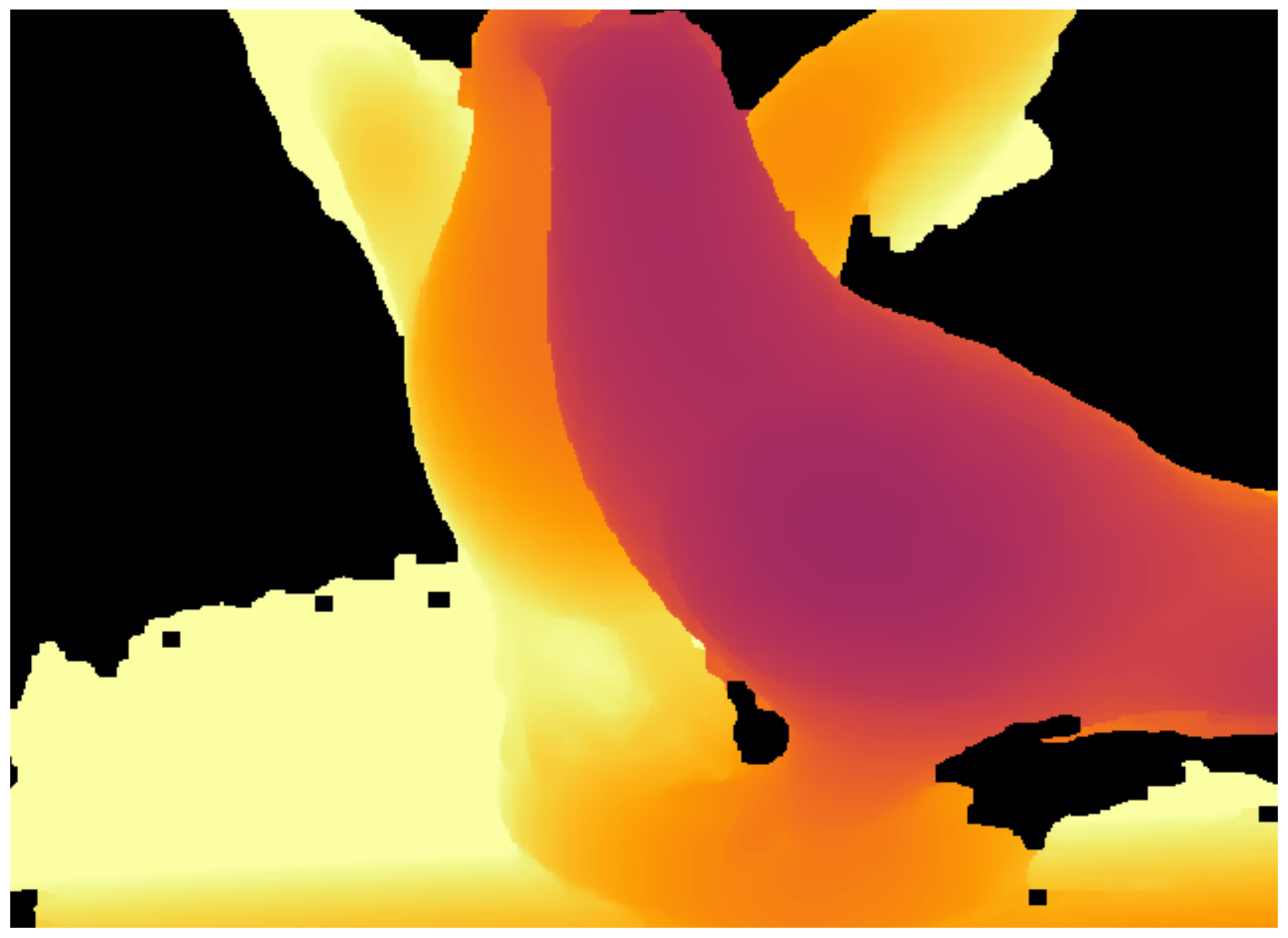}
        \caption[Ground Truth]%
        {{\small Ground Truth}}
    \end{subfigure}
    \hfill
    \begin{subfigure}[b]{0.30\linewidth}
    	\centering
        \includegraphics[width=\textwidth]{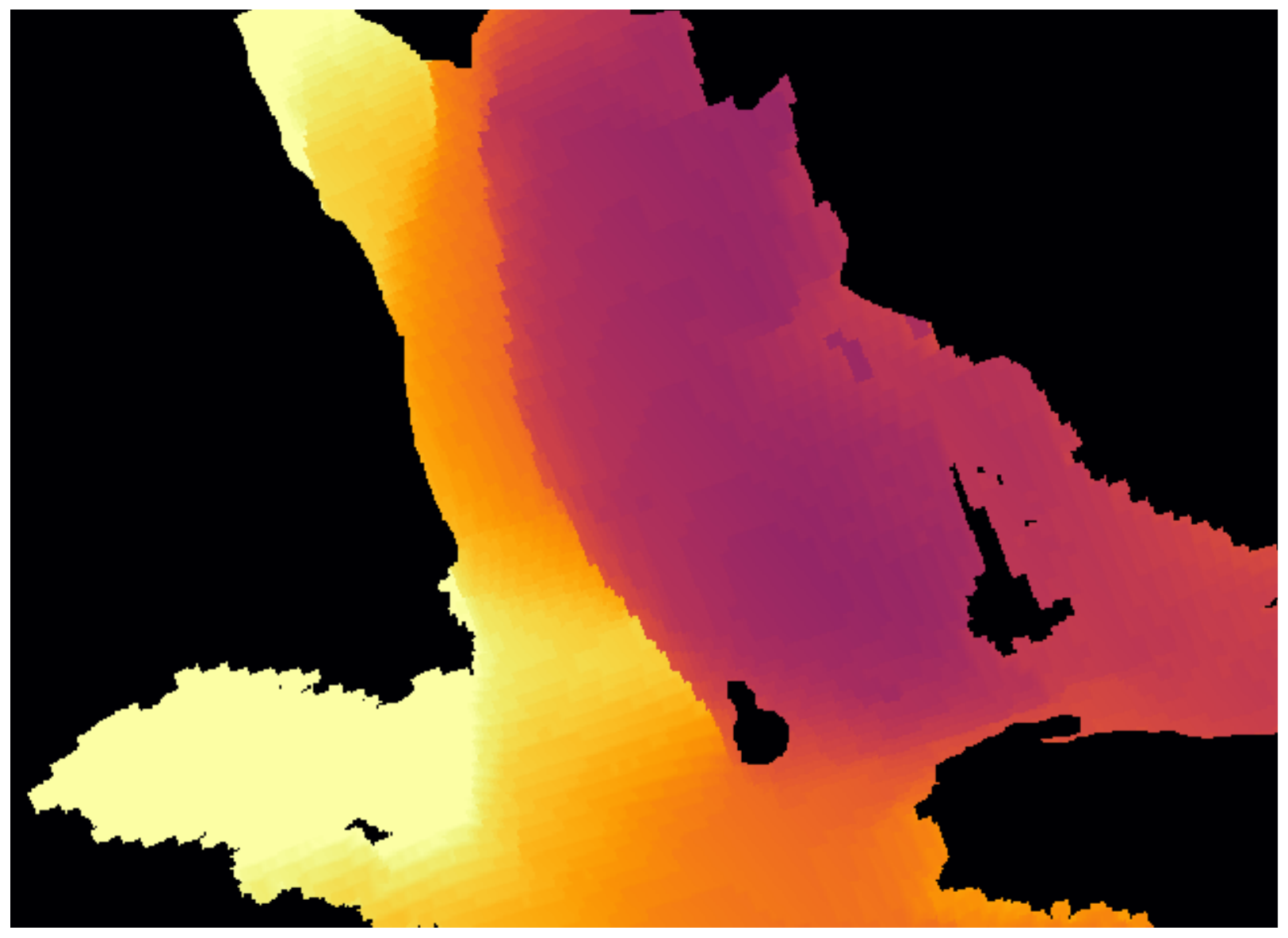}
        \caption[SurfaceNet \cite{Ji2017ICCV}]%
        {{\small SurfaceNet \cite{Ji2017ICCV}}}    
    \end{subfigure}
    \hfill
    \begin{subfigure}[b]{0.30\linewidth}
    	\centering
        \includegraphics[width=\textwidth]{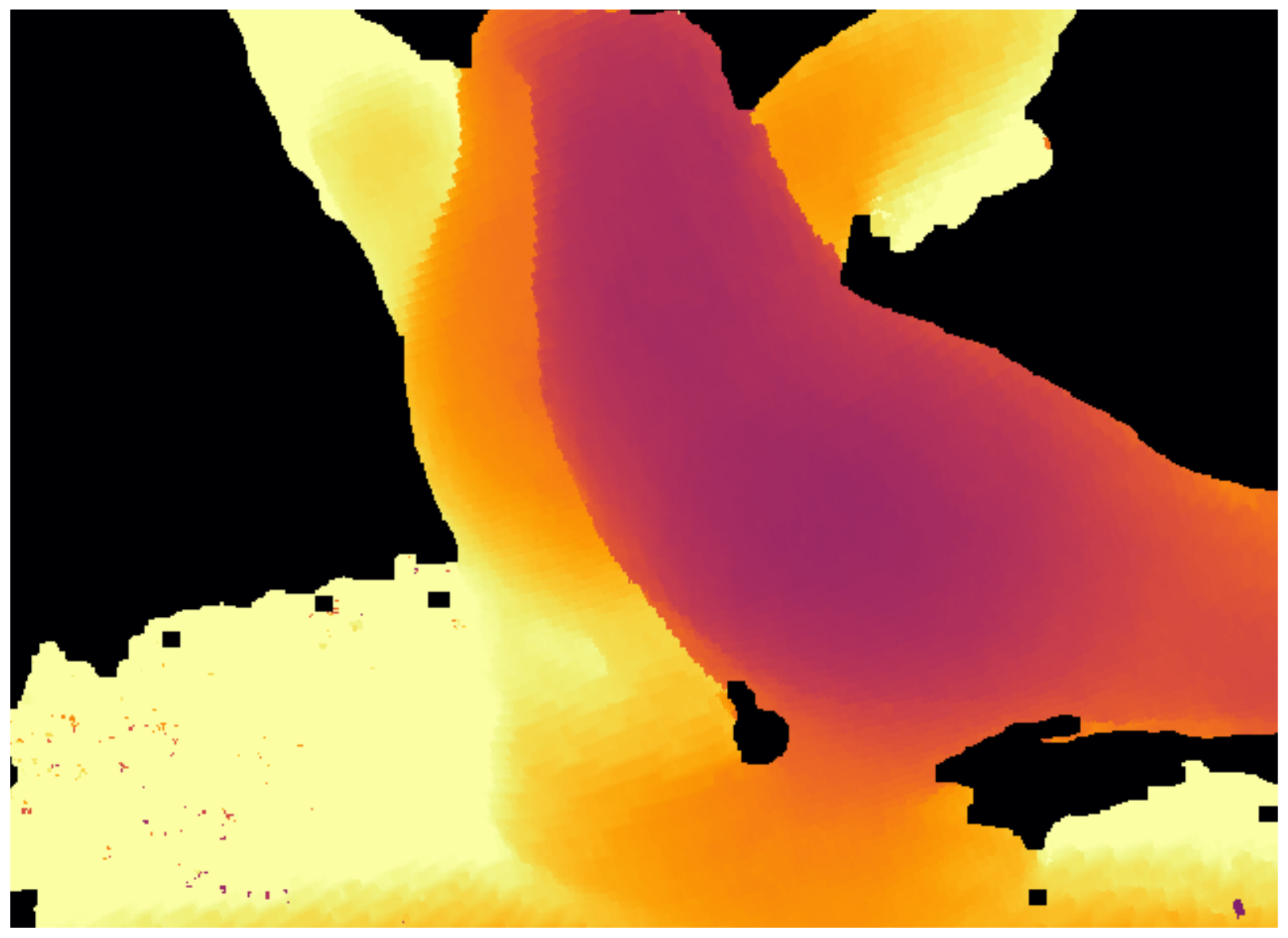}
        \caption[RayNet]%
        {{\small RayNet}}    
    \end{subfigure}
    \caption{{\bf Qualitative Results on DTU Dataset.} We visualize the ground-truth (a,d,g,j,m) and the depth reconstuctions using SurfaceNet (LR) (b,e,h,k,n) and RayNet (c,f,i,l,o).}
    \label{fig:budha_evaluation}
    \vspace{-1.25em}
 \end{figure}

 \figref{fig:budha_evaluation} shows a qualitative comparison between SurfaceNet (LR) and RayNet. It can be clearly observed that SurfaceNet often cannot reconstruct large parts of the object (\eg, the wings in the BIRD scene or part of the head in the BUDDHA scene) which it considers as unreliable. In contrast, RayNet generates more complete 3D reconstructions. RayNet is also significantly better at capturing object boundaries compared to SurfaceNet. 
 
\begin{table}[t!]
    \centering
    \resizebox{\columnwidth}{!}{
    \begin{tabular}{|l|C{1.0cm}|C{1.0cm}|C{1.0cm}|C{1.0cm}|}
        \hline
        Methods & \multicolumn{2}{c|}{Accuracy} &
        \multicolumn{2}{c|}{Completeness} \\
        \hline
        & Mean & Median & Mean & Median\\
        \hline
        {\bf BUDDHA} & & & &\\
        \hline
		ZNCC & 6.107 & 4.613 & 0.646 & 0.466\\
		SAD & 6.683 & 5.270 & 0.753 & 0.510\\
		SurfaceNet (HD) \cite{Ji2017ICCV} & 0.738 & 0.574 & 0.677 & 0.505\\
		SurfaceNet (LR) \cite{Ji2017ICCV} & 2.034 & 1.676 & 1.453 & 1.141\\
		Hartmann et al. \cite{Hartmann2017ICCV} & \textbf{0.637} & \textbf{0.206} & 1.057 & 0.475\\
		Ulusoy et al. \cite{Ulusoy2015THREEDV} & 4.784 & 3.522 & 0.953 & 0.402\\
		RayNet & 1.993 & 1.119 & \textbf{0.481} & \textbf{0.357}\\
        \hline
        {\bf BIRD} & & & &\\
        \hline
		ZNCC & 6.882 & 5.553 & \textbf{0.918} & 0.662\\
		SAD & 7.138 & 5.750 & 0.916 & \textbf{0.646}\\
		SurfaceNet (HD) \cite{Ji2017ICCV} & \textbf{1.493} & 1.249 & 1.612 & 0.888\\
		SurfaceNet (LR) \cite{Ji2017ICCV} & 2.887 & 2.468 & 2.330 & 1.556\\
		Hartmann et al. \cite{Hartmann2017ICCV} & 1.881 & \textbf{0.271} & 4.167 & 1.044\\
		Ulusoy et al. \cite{Ulusoy2015THREEDV} & 6.024 & 4.623 & 2.966 & 0.898\\
		RayNet & 2.618 & 1.680 & 0.983 & 0.668\\
        \hline
    \end{tabular}}
    \caption{{\bf Quantitative Results on DTU Dataset.} We present the mean and the median of accuracy and completeness measures for RayNet and various baselines. All results are reported in \emph{millimeters}, the default unit for DTU.}
    \label{tab:experimental_setup_2}
\end{table}

\subsection{Computational requirements}

The runtime and memory requirements of RayNet depend mainly on three factors: the number of voxels, the number of input images and the number of pixels/rays per image, which is typically equal to the image resolution.
All our experiments were computed on an Intel i7 computer with an Nvidia GTX Titan X GPU. We train RayNet end-to-end, which takes roughly $1$ day and requires $7$~GB per mini-batch update for the DTU dataset. Once the network is trained, it takes approximately $25$ minutes to obtain a full reconstruction of a typical scene from the DTU dataset. In contrast,  SurfaceNet (HD)~\cite{Ji2017ICCV} requires more than $4$ hours for this task. SurfaceNet (LR), which operates at the same voxel resolution as RayNet, requires $3$ hours. 

\section{Conclusion}
\label{sec:conclusion}

We propose RayNet, which is an end-to-end trained network that incorporates a CNN that learns multi-view image similarity with an MRF with ray potentials that explicitly models perspective projection and enforces occlusion constraints across viewpoints. We directly embed the physics of multi-view geometry into RayNet. Hence, RayNet is not required to learn these complex relationships from data. Instead, the network can  focus on learning view-invariant feature representations that are very difficult to model. Our experiments indicate that RayNet improves over learning-based approaches that do not incorporate multi-view geometry constraints and over model-based approaches that do not learn multi-view image matching.

Our current implementation precludes training models with a higher resolution than $256^3$ voxels.
While this resolution is finer than most existing learning-based approaches, it does not allow capturing high resolution details in large scenes.
In future work, we plan to adapt our method to higher resolution outputs using octree-based representations~\cite{Tatarchenko2017ICCV,Haene2017ARXIV,Ulusoy2015THREEDV}.
We also would like to extend RayNet to predict a semantic label per voxel in addition to occupancy. Recent works show such a joint prediction improves over reconstruction or segmentation in isolation~\cite{Savinov2015CVPR,Song2017CVPR}.

\section*{Acknowledgments}

This research was supported by the Max Planck ETH Center for Learning Systems.

{\small
	\bibliographystyle{ieee}
	\bibliography{bibliography_long,bibliography,bibliography_custom}

\begin{thebibliography}{10}\itemsep=-1pt

\bibitem{Aanes2016IJCV}
H.~Aan{\ae}s, R.~R. Jensen, G.~Vogiatzis, E.~Tola, and A.~B. Dahl.
\newblock Large-scale data for multiple-view stereopsis.
\newblock {\em International Journal of Computer Vision (IJCV)},
  120(2):153--168, 2016.

\bibitem{Abadi2016ARXIV}
M.~Abadi, P.~Barham, J.~Chen, Z.~Chen, A.~Davis, J.~Dean, M.~Devin,
  S.~Ghemawat, G.~Irving, M.~Isard, M.~Kudlur, J.~Levenberg, R.~Monga,
  S.~Moore, D.~G. Murray, B.~Steiner, P.~A. Tucker, V.~Vasudevan, P.~Warden,
  M.~Wicke, Y.~Yu, and X.~Zhang.
\newblock Tensorflow: {A} system for large-scale machine learning.
\newblock {\em arXiv.org}, 1605.08695, 2016.

\bibitem{Achlioptas2017ARXIV}
P.~Achlioptas, O.~Diamanti, I.~Mitliagkas, and L.~J. Guibas.
\newblock Representation learning and adversarial generation of 3d point
  clouds.
\newblock {\em arXiv.org}, 1707.02392, 2017.

\bibitem{Chang2015ARXIV}
A.~X. Chang, T.~A. Funkhouser, L.~J. Guibas, P.~Hanrahan, Q.~Huang, Z.~Li,
  S.~Savarese, M.~Savva, S.~Song, H.~Su, J.~Xiao, L.~Yi, and F.~Yu.
\newblock Shapenet: An information-rich 3d model repository.
\newblock {\em arXiv.org}, 1512.03012, 2015.

\bibitem{Choy2016ECCV}
C.~B. Choy, D.~Xu, J.~Gwak, K.~Chen, and S.~Savarese.
\newblock 3d-r2n2: {A} unified approach for single and multi-view 3d object
  reconstruction.
\newblock In {\em Proc. of the European Conf. on Computer Vision (ECCV)}, 2016.

\bibitem{Fan2017CVPR}
H.~Fan, H.~Su, and L.~J. Guibas.
\newblock A point set generation network for 3d object reconstruction from a
  single image.
\newblock {\em Proc. IEEE Conf. on Computer Vision and Pattern Recognition
  (CVPR)}, 2017.

\bibitem{Furukawa2009ICCV}
Y.~Furukawa, B.~Curless, S.~M. Seitz, and R.~Szeliski.
\newblock Reconstructing building interiors from images.
\newblock In {\em Proc. of the IEEE International Conf. on Computer Vision
  (ICCV)}, 2009.

\bibitem{Girdhar2016ECCV}
R.~Girdhar, D.~F. Fouhey, M.~Rodriguez, and A.~Gupta.
\newblock Learning a predictable and generative vector representation for
  objects.
\newblock In {\em Proc. of the European Conf. on Computer Vision (ECCV)}, 2016.

\bibitem{Gwak2017ARXIV}
J.~Gwak, C.~B. Choy, A.~Garg, M.~Chandraker, and S.~Savarese.
\newblock Weakly supervised generative adversarial networks for 3d
  reconstruction.
\newblock {\em arXiv.org}, 1705.10904, 2017.

\bibitem{Guney2016ACCV}
F.~Güney and A.~Geiger.
\newblock Deep discrete flow.
\newblock In {\em Proc. of the Asian Conf. on Computer Vision (ACCV)}, 2016.

\bibitem{Haene2014THREEDV}
C.~H{\"{a}}ne, L.~Heng, G.~H. Lee, A.~Sizov, and M.~Pollefeys.
\newblock Real-time direct dense matching on fisheye images using
  plane-sweeping stereo.
\newblock In {\em Proc. of the International Conf. on 3D Vision (3DV)}, 2014.

\bibitem{Haene2017ARXIV}
C.~H{\"{a}}ne, S.~Tulsiani, and J.~Malik.
\newblock Hierarchical surface prediction for 3d object reconstruction.
\newblock {\em arXiv.org}, 1704.00710, 2017.

\bibitem{Hartley2004}
R.~I. Hartley and A.~Zisserman.
\newblock {\em Multiple View Geometry in Computer Vision}.
\newblock Cambridge University Press, second edition, 2004.

\bibitem{Hartmann2017ICCV}
W.~Hartmann, S.~Galliani, M.~Havlena, L.~{Van Gool}, and K.~Schindler.
\newblock Learned multi-patch similarity.
\newblock In {\em Proc. of the IEEE International Conf. on Computer Vision
  (ICCV)}, 2017.

\bibitem{Ilg2017CVPR}
E.~Ilg, N.~Mayer, T.~Saikia, M.~Keuper, A.~Dosovitskiy, and T.~Brox.
\newblock Flownet 2.0: Evolution of optical flow estimation with deep networks.
\newblock {\em Proc. IEEE Conf. on Computer Vision and Pattern Recognition
  (CVPR)}, 2017.

\bibitem{Ji2017ICCV}
M.~Ji, J.~Gall, H.~Zheng, Y.~Liu, and L.~Fang.
\newblock {SurfaceNet:} an end-to-end 3d neural network for multiview
  stereopsis.
\newblock In {\em Proc. of the IEEE International Conf. on Computer Vision
  (ICCV)}, 2017.

\bibitem{Kar2017NIPS}
A.~Kar, C.~H{\"{a}}ne, and J.~Malik.
\newblock Learning a multi-view stereo machine.
\newblock In {\em Advances in Neural Information Processing Systems (NIPS)},
  2017.

\bibitem{Kingma2015ICLR}
D.~P. Kingma and J.~Ba.
\newblock Adam: {A} method for stochastic optimization.
\newblock In {\em Proc. of the International Conf. on Learning Representations
  (ICLR)}, 2015.

\bibitem{Liu2014PAMI}
S.~Liu and D.~Cooper.
\newblock Statistical inverse ray tracing for image-based 3d modeling.
\newblock {\em IEEE Trans. on Pattern Analysis and Machine Intelligence
  (PAMI)}, 36(10):2074--2088, 2014.

\bibitem{Luo2016CVPR}
W.~Luo, A.~Schwing, and R.~Urtasun.
\newblock Efficient deep learning for stereo matching.
\newblock In {\em Proc. IEEE Conf. on Computer Vision and Pattern Recognition
  (CVPR)}, 2016.

\bibitem{Mayer2016CVPR}
N.~Mayer, E.~Ilg, P.~Haeusser, P.~Fischer, D.~Cremers, A.~Dosovitskiy, and
  T.~Brox.
\newblock A large dataset to train convolutional networks for disparity,
  optical flow, and scene flow estimation.
\newblock In {\em Proc. IEEE Conf. on Computer Vision and Pattern Recognition
  (CVPR)}, 2016.

\bibitem{Nowozin2011FTCGV}
S.~Nowozin and C.~H. Lampert.
\newblock Structured learning and prediction in computer vision.
\newblock {\em Foundations and Trends in Computer Graphics and Vision}, 2011.

\bibitem{Pollard2007CVPR}
T.~Pollard and J.~L. Mundy.
\newblock Change detection in a 3-d world.
\newblock In {\em Proc. IEEE Conf. on Computer Vision and Pattern Recognition
  (CVPR)}, 2007.

\bibitem{Ranjan2017CVPR}
A.~Ranjan and M.~Black.
\newblock Optical flow estimation using a spatial pyramid network.
\newblock In {\em Proc. IEEE Conf. on Computer Vision and Pattern Recognition
  (CVPR)}, 2017.

\bibitem{Restrepo2014JPRS}
M.~I. Restrepo, A.~O. Ulusoy, and J.~L. Mundy.
\newblock Evaluation of feature-based 3-d registration of probabilistic
  volumetric scenes.
\newblock {\em ISPRS Journal of Photogrammetry and Remote Sensing (JPRS)},
  2014.

\bibitem{Rezende2016NIPS}
D.~J. Rezende, S.~M.~A. Eslami, S.~Mohamed, P.~Battaglia, M.~Jaderberg, and
  N.~Heess.
\newblock Unsupervised learning of 3d structure from images.
\newblock In {\em Advances in Neural Information Processing Systems (NIPS)},
  2016.

\bibitem{Riegler2017THREEDV}
G.~Riegler, A.~O. Ulusoy, H.~Bischof, and A.~Geiger.
\newblock {OctNetFusion}: Learning depth fusion from data.
\newblock In {\em Proc. of the International Conf. on 3D Vision (3DV)}, 2017.

\bibitem{Savinov2015CVPR}
N.~Savinov, L.~Ladicky, C.~Häne, and M.~Pollefeys.
\newblock Discrete optimization of ray potentials for semantic 3d
  reconstruction.
\newblock In {\em Proc. IEEE Conf. on Computer Vision and Pattern Recognition
  (CVPR)}, 2015.

\bibitem{Seitz2006CVPR}
S.~M. Seitz, B.~Curless, J.~Diebel, D.~Scharstein, and R.~Szeliski.
\newblock A comparison and evaluation of multi-view stereo reconstruction
  algorithms.
\newblock In {\em Proc. IEEE Conf. on Computer Vision and Pattern Recognition
  (CVPR)}, 2006.

\bibitem{Song2017CVPR}
S.~Song, F.~Yu, A.~Zeng, A.~X. Chang, M.~Savva, and T.~Funkhouser.
\newblock Semantic scene completion from a single depth image.
\newblock In {\em Proc. IEEE Conf. on Computer Vision and Pattern Recognition
  (CVPR)}, 2017.

\bibitem{Tatarchenko2017ICCV}
M.~Tatarchenko, A.~Dosovitskiy, and T.~Brox.
\newblock Octree generating networks: Efficient convolutional architectures for
  high-resolution 3d outputs.
\newblock In {\em Proc. of the IEEE International Conf. on Computer Vision
  (ICCV)}, 2017.

\bibitem{Tulsiani2017CVPR}
S.~Tulsiani, T.~Zhou, A.~A. Efros, and J.~Malik.
\newblock Multi-view supervision for single-view reconstruction via
  differentiable ray consistency.
\newblock In {\em Proc. IEEE Conf. on Computer Vision and Pattern Recognition
  (CVPR)}, 2017.

\bibitem{Ulusoy2016CVPR}
A.~O. Ulusoy, M.~Black, and A.~Geiger.
\newblock Patches, planes and probabilities: A non-local prior for volumetric
  3d reconstruction.
\newblock In {\em Proc. IEEE Conf. on Computer Vision and Pattern Recognition
  (CVPR)}, 2016.

\bibitem{Ulusoy2017CVPR}
A.~O. Ulusoy, M.~Black, and A.~Geiger.
\newblock Semantic multi-view stereo: Jointly estimating objects and voxels.
\newblock In {\em Proc. IEEE Conf. on Computer Vision and Pattern Recognition
  (CVPR)}, 2017.

\bibitem{Ulusoy2015THREEDV}
A.~O. Ulusoy, A.~Geiger, and M.~J. Black.
\newblock Towards probabilistic volumetric reconstruction using ray potentials.
\newblock In {\em Proc. of the International Conf. on 3D Vision (3DV)}, 2015.

\bibitem{Wu2011CVPR}
C.~Wu, S.~Agarwal, B.~Curless, and S.~Seitz.
\newblock Multicore bundle adjustment.
\newblock In {\em Proc. IEEE Conf. on Computer Vision and Pattern Recognition
  (CVPR)}, 2011.

\bibitem{Wu2016NIPS}
J.~Wu, C.~Zhang, T.~Xue, B.~Freeman, and J.~Tenenbaum.
\newblock Learning a probabilistic latent space of object shapes via 3d
  generative-adversarial modeling.
\newblock In {\em Advances in Neural Information Processing Systems (NIPS)},
  2016.

\bibitem{Wu2015CVPR}
Z.~Wu, S.~Song, A.~Khosla, F.~Yu, L.~Zhang, X.~Tang, and J.~Xiao.
\newblock 3d shapenets: {A} deep representation for volumetric shapes.
\newblock In {\em Proc. IEEE Conf. on Computer Vision and Pattern Recognition
  (CVPR)}, 2015.

\bibitem{Yan2016NIPS}
X.~Yan, J.~Yang, E.~Yumer, Y.~Guo, and H.~Lee.
\newblock Perspective transformer nets: Learning single-view 3d object
  reconstruction without 3d supervision.
\newblock In {\em Advances in Neural Information Processing Systems (NIPS)},
  2016.

\bibitem{Zagoruyko2015CVPR}
S.~Zagoruyko and N.~Komodakis.
\newblock Learning to compare image patches via convolutional neural networks.
\newblock In {\em Proc. IEEE Conf. on Computer Vision and Pattern Recognition
  (CVPR)}, 2015.

\bibitem{Zbontar2014ARXIV}
J.~Zbontar and Y.~LeCun.
\newblock Computing the stereo matching cost with a convolutional neural
  network.
\newblock {\em arXiv.org}, 1409.4326, 2014.

\bibitem{Zbontar2016JMLR}
J.~{\v{Z}}bontar and Y.~LeCun.
\newblock Stereo matching by training a convolutional neural network to compare
  image patches.
\newblock {\em Journal of Machine Learning Research (JMLR)}, 17(65):1--32,
  2016.

\end{thebibliography}
}

\end{document}